\documentclass[11pt, a4paper, logo, copyright]{googlecloud}

\pdftrailerid{redacted}

\makeatletter
\renewcommand\bibentry[1]{\nocite{#1}{\frenchspacing\@nameuse{BR@r@#1\@extra@b@citeb}}}
\makeatother

\usepackage{kantlipsum, lipsum}
\usepackage{dsfont}

\usepackage[authoryear, sort&compress, round]{natbib}

\usepackage[utf8]{inputenc} 
\usepackage[T1]{fontenc}    
\usepackage{url}            
\usepackage{booktabs}       
\usepackage{nicefrac}       
\usepackage{microtype}      
\usepackage{amsmath}
\usepackage{graphicx}
\usepackage{multicol}
\usepackage{hyperref}       
\usepackage[nameinlink]{cleveref}
\crefname{appendix}{appendix}{appendices}
\Crefname{appendix}{Appendix}{Appendices}

\usepackage{bbm}
\usepackage{multirow}
\usepackage{soul}
\usepackage{float}
\usepackage{wrapfig}
\usepackage{blindtext}
\usepackage{tablefootnote}
\usepackage{amsfonts}
\usepackage[flushleft]{threeparttable}
\usepackage{bbding}
\usepackage{xcolor}
\usepackage{xspace}
\makeatletter
\DeclareRobustCommand\onedot{\futurelet\@let@token\@onedot}
\def\@onedot{\ifx\@let@token.\else.\null\fi\xspace}
\def\eg{\emph{e.g}\onedot}

\usepackage{bm}
\usepackage{arydshln}
\usepackage{enumitem}
\usepackage{setspace}
\usepackage{color}
\usepackage{algorithm}
\usepackage{longtable}

\usepackage[normalem]{ulem}
\usepackage{ulem}
\usepackage[nomargin,inline,marginclue,draft]{fixme}
\usepackage{balance}
\usepackage{verbatim}
\usepackage{diagbox}
\usepackage{changepage}
\usepackage{amssymb}
\usepackage{pifont}
\usepackage{algpseudocode}

\usepackage{listings}
\usepackage{tcolorbox}
\tcbuselibrary{listings, breakable, skins}
\usepackage[table]{xcolor}

\usepackage{subcaption} 
\usepackage{mathtools}
\usepackage{amsthm}
\usepackage{tikz}
\usepackage[disable,textsize=tiny]{todonotes}

\usepackage{graphicx}   
\usepackage{subcaption} 
\usepackage{booktabs}   
\usepackage{multirow}   
\usepackage{array}

\usepackage{listings}
\usepackage{etoolbox}

\lstdefinestyle{pythonstyle}{
    language=Python,
    basicstyle=\ttfamily\small,
    keywordstyle=\color{blue},
    commentstyle=\color{gray},
    stringstyle=\color{orange},
    numbers=none, 
    stepnumber=1,
    numbersep=5pt,
    backgroundcolor=\color{white},
    frame=single,
    breaklines=true, 
    breakatwhitespace=true,
    captionpos=b,
    xleftmargin=0.5em, 
    xrightmargin=0.5em, 
    showstringspaces=false, 
    columns=flexible, 
}

\crefname{lstlisting}{algorithm}{algorithms}
\Crefname{lstlisting}{Algorithm}{Algorithms}

\definecolor{bananayellow}{HTML}{FFF0C9}
\newtcolorbox{promptbox}[2][blue]{
  colback=white,
  colframe=#1,
  colbacktitle=#1!10,
  coltitle=black,
  title=#2,
  fonttitle=\bfseries\sffamily,
  breakable,
  enhanced,
  boxrule=0.8pt
}
\theoremstyle{plain}

\theoremstyle{definition}

\theoremstyle{remark}

\def\ourmodel{VGGRPO}

\usepackage{array}
\definecolor{row}{RGB}{235, 245, 251}

\newcommand{\projecturl}{https://zhaochongan.github.io/projects/VGGRPO}
\newcommand{\projectpage}{\href{https://zhaochongan.github.io/projects/VGGRPO}{\textcolor{magenta}{Project Page}}\xspace}

\begin{document}

\title{VGGRPO: Towards World-Consistent Video Generation with 4D Latent Reward}

\correspondingauthor{zhan@di.ku.dk}

\author[1 2 *]{Zhaochong An}
\author[1,3]{Orest Kupyn}
\author[1,4]{Théo Uscidda}
\author[1]{Andrea Colaco}
\author[1]{Karan Ahuja}
\author[2]{Serge Belongie}
\author[1]{Mar Gonzalez-Franco}
\author[1]{Marta Tintore Gazulla}

\affil[1]{Google}
\affil[2]{University of Copenhagen}
\affil[3]{University of Oxford}
\affil[4]{CREST-ENSAE, Institut Polytechnique de Paris}

\begin{abstract}
Large-scale video diffusion models achieve impressive visual quality, yet often fail to preserve geometric consistency. 
Prior approaches improve consistency either by augmenting the generator with additional modules or applying geometry-aware alignment. 
However, architectural modifications can compromise the generalization of internet-scale pretrained models, while existing alignment methods are limited to static scenes and rely on RGB-space rewards that require repeated VAE decoding, incurring substantial compute overhead and failing to generalize to highly dynamic real-world scenes.
To preserve the pretrained capacity while improving geometric consistency, we propose \textbf{VGGRPO} (\textbf{V}isual \textbf{G}eometry \textbf{GRPO}), a latent geometry-guided framework for geometry-aware video post-training. 
VGGRPO introduces a \emph{Latent Geometry Model} (LGM) that stitches video diffusion latents to geometry foundation models, enabling direct decoding of scene geometry from the latent space. 
By constructing LGM from a geometry model with 4D reconstruction capability, VGGRPO naturally extends to dynamic scenes, overcoming the static-scene limitations of prior methods. 
Building on this, we perform latent-space Group Relative Policy Optimization with two complementary rewards: a \emph{camera motion smoothness} reward that penalizes jittery trajectories, and a \emph{geometry reprojection consistency} reward that enforces cross-view geometric coherence. 
Experiments on both static and dynamic benchmarks show that VGGRPO improves camera stability, geometry consistency, and overall quality while eliminating costly VAE decoding, making latent-space geometry-guided reinforcement an efficient and flexible approach to world-consistent video generation.
\end{abstract}

\maketitle

\begin{figure*}[!t]
    \centering
    \includegraphics[width=\textwidth]{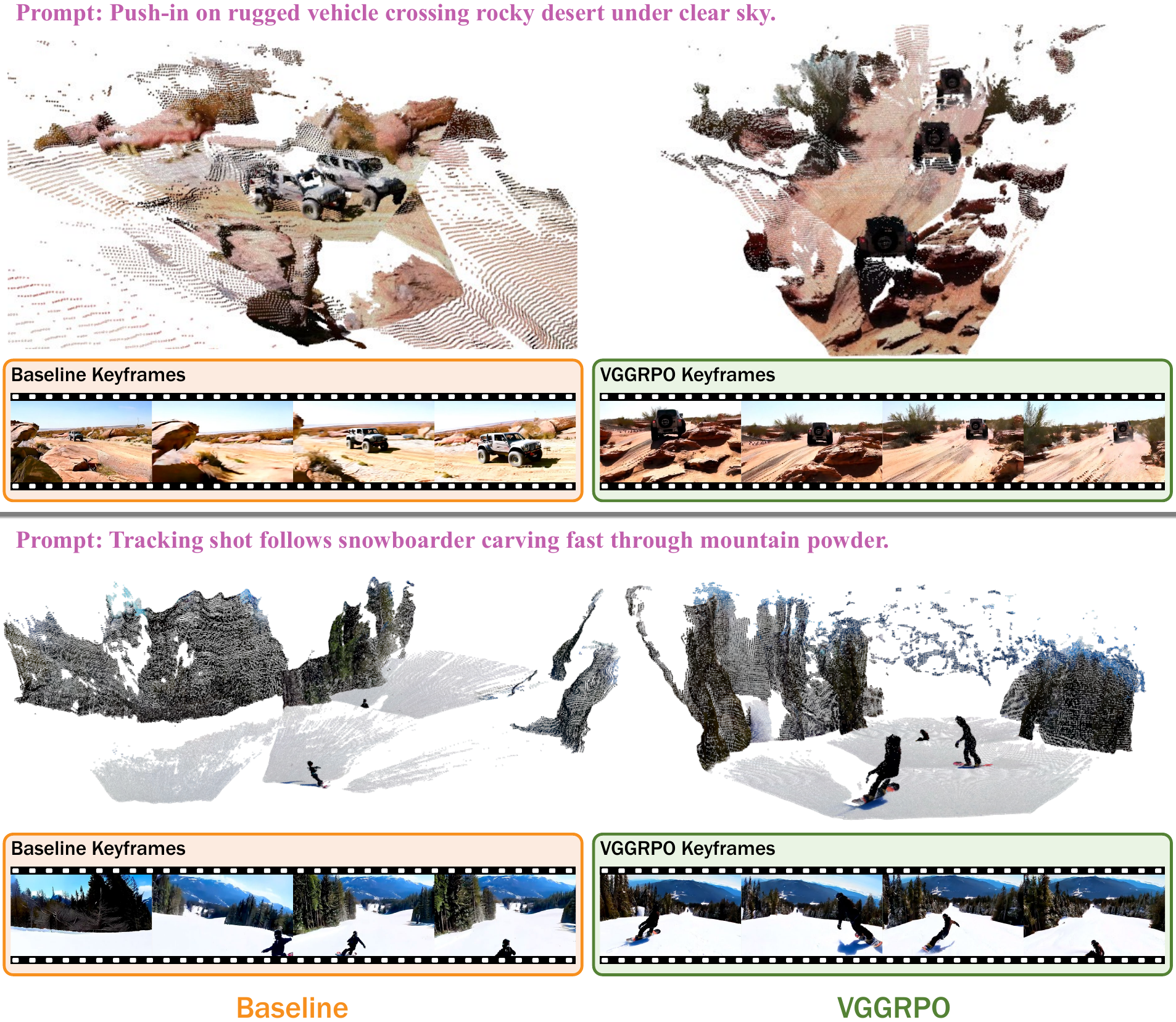}
    \caption{\textbf{World-consistent Video Generations with VGGRPO.}
We compare the baseline video diffusion model (\textit{left}, orange) with the VGGRPO-aligned model (\textit{right}, green).
Each example depicts a challenging dynamic scene; we visualize representative keyframes from the generated video and reconstructed scene geometry from the inferred 4D scene representation.
VGGRPO produces markedly more coherent scene structure and smoother camera motion over time, reducing geometric drift and structural artifacts in challenging dynamic settings.}
    \label{fig:teaser}
\end{figure*}

\section{Introduction}
Recent video diffusion models~\citep{veo,zhou2025scaling,qiu2025histream,liu2025tuna,an2025onestory} have achieved impressive visual fidelity and broad generalization by training on large volumes of diverse, high-quality data. 
However, they often lack 3D and motion consistency~\citep{park2025steerx,wang2026worldcompass,bhowmik2025moalign,xue2025mogan,gao2026pulse,wang2026chain}, exhibiting geometric drift, unstable camera trajectory, and inconsistent scene structure. 
These issues are critical for downstream applications~\citep{an2026video,gao2026dreamdojo,le2025gravity,jiang2026wovr,ren2026videoworld,an2024multimodality,yuan2026fast,liu2026ir3d,agarwal2026cosmos,intelligence2026pi} such as embodied AI and physics-aware simulation, where stable camera motion and coherent 3D geometry are required. 

To mitigate these issues, existing efforts largely follow two paradigms. 
The first paradigm \textit{injects geometric structure into the generator} via additional conditioning modules~\citep{yu2024viewcrafter,ren2025gen3c,cao2025uni3c} or extra loss components~\citep{geometry_forcing,ViCoDR}. 
For example, point cloud-conditioned diffusion models~\citep{ren2025gen3c, cao2025uni3c} impose pixel-wise constraints from 3D inputs to improve static-scene consistency, while other approaches~\citep{wvd,bai2025geovideo,huang2025jog3r,dai2025fantasyworld} augment video diffusion with auxiliary geometry prediction to improve scene generation. 
While effective, these modifications often increase architectural and computational complexity and can constrain the model, weakening the broad generalization inherited from large-scale pretraining.
The second paradigm performs \textit{post-training alignment} inspired by reinforcement learning. 
Recent approaches adapt Direct Preference Optimization~\citep{dpo,liuimproving} and compute rewards from sparse epipolar constraints~\citep{kupyn2025epipolar} or dense correspondences~\citep{du2026videogpa,gu2025geco} predicted by external geometry models~\citep{wang2025vggt}. 
However, these methods rely on offline preference data collection, yielding off-policy optimization. 
Moreover, rewards are typically evaluated in pixel space, requiring repeated VAE decoding, which significantly increases compute and memory overhead; RGB-based rewards are also sensitive to decoding noise and low-level pixel variations~\citep{mi2025video,gotext}, further weakening the optimization signals.
Finally, these geometric reward formulations are limited to static scenes, as their underlying assumptions~\citep{kupyn2025epipolar} and correspondence pipelines~\citep{du2026videogpa,gu2025geco} do not extend to complex dynamic videos.

In parallel, recent geometry foundation models~\citep{wang2025vggt,karhade2025any4d} have demonstrated that feed-forward networks can recover dense geometry and camera motion from static and dynamic image sequences, encoding strong geometric priors learned at scale.
This raises a key question: \emph{can we leverage these geometry priors while avoiding the cost and instability of RGB-based reward evaluation?}

We introduce \textbf{VGGRPO} (\textbf{V}isual \textbf{G}eometry \textbf{GRPO}), a latent geometry-guided, group-based reinforcement learning framework for video post-training. 
VGGRPO comprises two tightly coupled components. 
First, we construct a \emph{Latent Geometry Model} (LGM) that connects video diffusion latents to a geometry foundation model via a lightweight stitching layer, thereby preserving its geometric priors.
By operating directly in the VAE latent space, LGM further enables efficient scene-geometry extraction from latents without RGB decoding.
Importantly, LGM is model-agnostic and can be instantiated with different geometry foundation models~\citep{wang2025vggt,karhade2025any4d}, enabling VGGRPO to benefit from ongoing progress in this domain. 
When connected to models supporting dynamic 4D reconstruction~\citep{karhade2025any4d}, LGM allows VGGRPO to support dynamic videos, extending beyond the static-scene assumptions of prior geometry-consistency methods~\citep{kupyn2025epipolar,du2026videogpa}.

Second, building on LGM, VGGRPO performs \emph{latent-space} Group Relative Policy Optimization without repeated VAE decoding, substantially reducing the cost of group-based updates. 
To optimize for temporally smooth camera motion and coherent 3D structure across viewpoints, we design two complementary rewards: a \emph{camera motion smoothness reward} that encourages stable camera trajectories, and a \emph{geometry reprojection consistency reward} that enforces cross-view geometric coherence. 
Together, these rewards improve camera stability and 3D consistency, resulting in more realistic, world-consistent video generation, as shown in~\Cref{fig:teaser}.

Extensive experiments show that VGGRPO yields consistent gains on both static- and dynamic-scene benchmarks across camera motion smoothness, geometric consistency, and overall video quality.
Compared to RGB-based alignment strategies, VGGRPO latent rewards efficiently incorporate geometry priors and support dynamic scenes, providing a practical solution for world-consistent video post-training.
Our contributions are summarized as follows:
\begin{itemize}
    \item We show that reliable geometry-driven rewards can be computed directly in latent space, enabling efficient video post-training without repeated RGB decoding.
    \item We propose a \textbf{Latent Geometry Model} that stitches diffusion latents to geometry foundation models via a lightweight connector, enabling extraction of geometric predictions from latents without pixel-space inputs.
    \item We introduce \textbf{VGGRPO}, a latent-space, group-based reinforcement learning framework with complementary camera-motion and geometry rewards that jointly improve camera smoothness and 3D consistency for world-consistent video generation.
\end{itemize}

\section{Related Work}
Based on how existing methods improve geometric consistency, we divide the literature review into geometrically consistent video generation and diffusion model alignment methods.

\subsection{Geometrically Consistent Video Generation}
Large-scale diffusion models~\citep{Sora, Runway2024Gen3, PikaLabs2024Pika, LumaLabs2024DreamMachine,hacohen2026ltx,li2026skyreels,huang2025jova} trained with rectified flow~\citep{rectified_flow} have significantly advanced video generation, yet often exhibit geometric drift that undermines scene realism. 
Geometrically and world-consistent video generation is crucial for downstream applications~\citep{gao2026dreamdojo,genie3,liang2025wonderland} requiring stable camera motion and coherent scene geometry.
Existing approaches to improve geometric consistency broadly fall into two paradigms.
\textit{(i) Architecture-level geometry integration.}
Point cloud-conditioned diffusion methods~\citep{ren2025gen3c, cao2025uni3c, yu2024viewcrafter,wang2025anchorweave,li2025vmem} introduces explicit 3D conditioning to improve static scene consistency. 
Other approaches augment diffusion models with auxiliary geometry prediction modules~\citep{hu2026geometry,zhang2025dualcamctrl}. 
World-consistent Video Diffusion~\citep{wvd} jointly models RGB and XYZ frames by treating 3D coordinates as an additional modality. 
GeoVideo~\citep{bai2025geovideo} incorporates depth prediction with cross-frame consistency losses, while FantasyWorld~\citep{dai2025fantasyworld} trains additional decoders to decode scene geometry alongside RGB frames. 
Although effective, these approaches increase architectural and computational complexity and are limited to static scenes. 
Furthermore, such modifications can restrict generative flexibility and weaken generalization.
\textit{(ii) Training-time regularization.}
Another direction introduces extra supervision during training without adding new modules. 
Geometry Forcing~\citep{geometry_forcing} aligns diffusion features with a foundational geometry model~\citep{wang2025vggt}, and ViCoDR~\citep{ViCoDR} incorporates 3D correspondence losses into video diffusion training. 
However, these methods typically require full model fine-tuning or training from scratch, which can compromise the broad generalization from large-scale pretraining.
In contrast, VGGRPO performs geometry-aware latent-space post-training with on-policy reward optimization, both improving geometric consistency and preserving generalization.
Importantly, it naturally extends to dynamic scenes, which prior methods largely do not address.

\subsection{Diffusion Model Alignment}
Large-scale diffusion models~\citep{rectified_flow} are trained to match broad data distributions, which may not align generations with task-specific objectives (\eg, aesthetics or physical constraints). 
Post-training alignment tackles this gap. 
Early work on image generation~\citep{sdxl, ldm} fine-tuned diffusion models on data filtered by aesthetic classifiers~\citep{Schuhmann2022LAION}. 
Later methods cast denoising as a sequential decision process: DDPO~\citep{ddpo} and DPOK~\citep{dpok} apply policy-gradient updates under distributional constraints, while Diffusion-DPO~\citep{diffdpo} adapts Direct Preference Optimization (DPO)~\citep{dpo} to diffusion models using pairwise preferences data. 
Flow-DPO~\citep{liuimproving} further extends DPO to rectified-flow models. 
Some works target physical accuracy: PISA~\citep{pisa} improves physical stability via multi-component rewards, and PhysCorr~\citep{wang2025physcorr} enhances physical realism through VLM-based rewards.
More recently, alignment has been explored for geometry-aware video generation~\citep{wang2026world}.
Epipolar-DPO~\citep{kupyn2025epipolar} incorporates epipolar constraints, and VideoGPA~\citep{du2026videogpa} extends this direction with dense geometry rewards~\citep{wang2025vggt}; however, they assume static-scene consistency, limiting applicability to dynamic videos.
Moreover, these approaches require computing rewards in pixel space and rely on offline preference data collection, incurring repeated RGB decoding and limiting optimization efficiency.
In parallel, Group Relative Policy Optimization (GRPO)~\citep{grpo} offers an on-policy alternative by sampling from the current model during training, keeping reward signals on-policy without a fixed preference dataset. 
Flow-GRPO~\citep{flowgrpo} and DanceGRPO~\citep{xue2025dancegrpo} adapt this framework to flow-based generators~\citep{rectified_flow}, but still require RGB decoding to evaluate rewards; designing effective latent-space rewards remains challenging. 
Motivated by these limitations, we propose VGGRPO, which performs GRPO-based geometry alignment directly in latent space via a latent geometry model, removing the RGB decoding bottleneck and enabling flexible geometry-aware alignment for dynamic scenes.

\section{Methodology}
\begin{figure}[t]
    \centering
    \includegraphics[width=\linewidth]{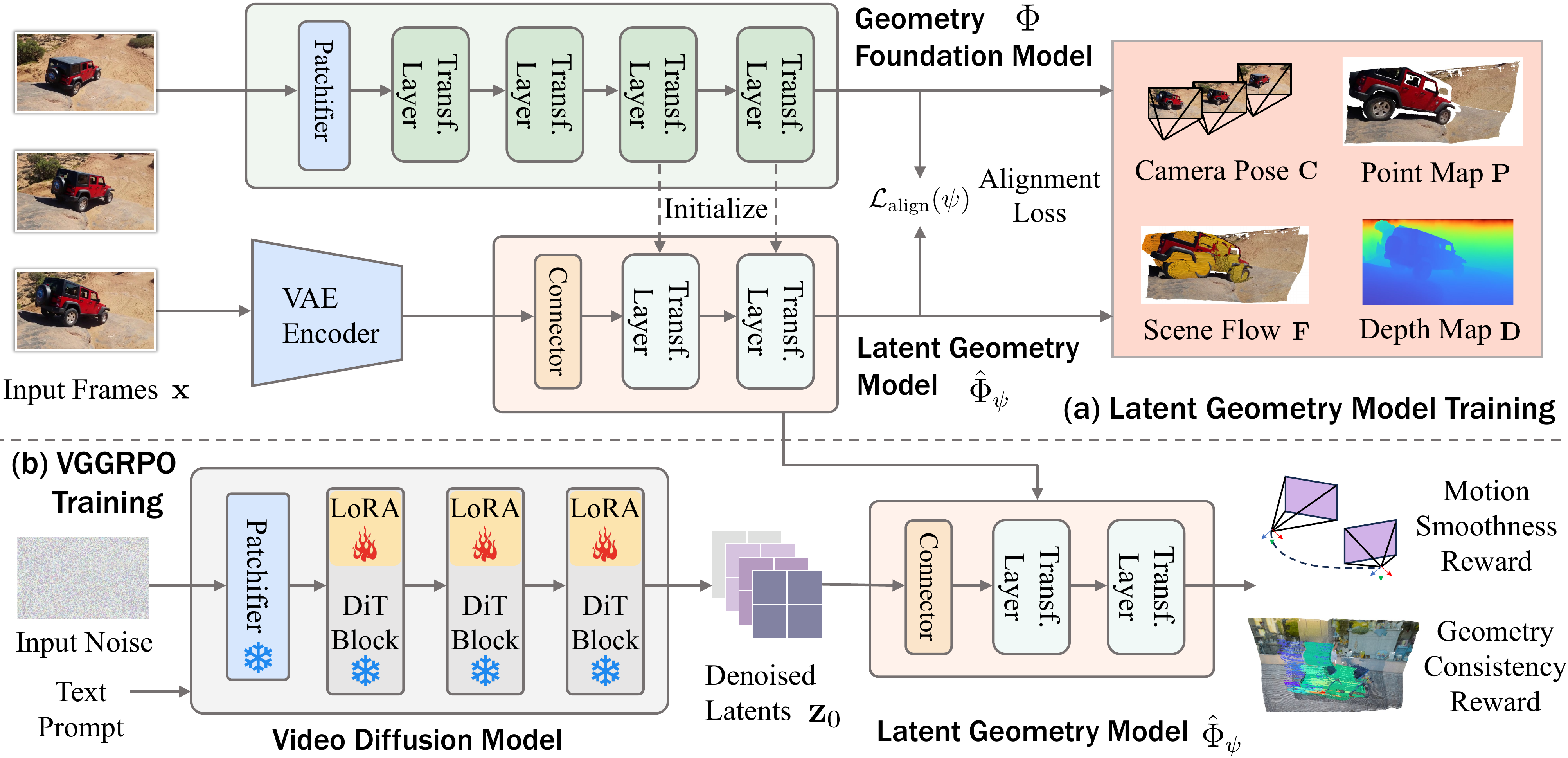}
\caption{\textbf{Method Overview.}
(a) \textbf{Latent Geometry Model.} We connect latents from the diffusion VAE encoder to a geometry foundation model via a lightweight connector, yielding a Latent Geometry Model that predicts 4D scene geometry directly from video latents.
(b) \textbf{VGGRPO training.} We perform latent-space GRPO using two complementary rewards, camera motion smoothness and geometry reprojection consistency, computed entirely in latent space with the latent geometry model. Together, these components align the video diffusion model toward 4D world-consistent generation on both static and dynamic scenes.}

\label{fig:main_fig}
\end{figure}

We now detail the formulation of~\ourmodel, which couples a latent geometry model with group-based reinforcement optimization for geometry-aware video post-training. 
As shown in~\Cref{fig:main_fig}, our method comprises two components: 
(1) a \emph{Latent Geometry Model} constructed via model stitching, which enables geometry extraction directly from diffusion latents without RGB decoding; and 
(2) \emph{\ourmodel~training}
which performs latent-space GRPO using two complementary rewards: \emph{camera motion smoothness} and \emph{geometry reprojection consistency}. These rewards jointly encourage stable camera trajectories and cross-view geometric coherence. 
We first introduce preliminaries in~\Cref{sec:preliminary}, then describe the latent geometry model in~\Cref{sec:stitching} and \ourmodel~training procedure in~\Cref{sec:latentgrpo}.

\subsection{Preliminaries}
\label{sec:preliminary}
\noindent\textbf{Flow-Based Group Relative Policy Optimization} formulates the denoising process of rectified flow models~\citep{rectified_flow} as a multi-step MDP~\citep{ddpo} and applies GRPO~\citep{grpo,flowgrpo,xue2025dancegrpo,li2025growing,zheng2025diffusionnft} with an ODE-to-SDE conversion for stochastic exploration.
This framework operates on RGB video frames $\mathbf{x} = \{\mathbf{I}_i\}_{i=1}^N$.
Let $\mathcal{P}$ be a set of text prompts and $\pi_\theta$ a policy parametrized by a velocity field $v_\theta(\mathbf{x}_t, t, p)$. The goal is to maximize the expected reward with KL regularization toward a reference policy $\pi_{\mathrm{ref}}$:
\begin{equation}
\label{eq:reward_maximization}
\max_\theta \, \mathbb{E}_{p\sim \mathcal{P}, \,\mathbf{x}_0 \sim \pi_\theta(\cdot\mid p)}[r(\mathbf{x}_0, p)] - D_\mathrm{KL}(\pi_\theta \|\,\pi_{\mathrm{ref}})
\end{equation}
To optimize this objective, GRPO draws $K$ samples from the current policy for each prompt, each consisting of a full denoising trajectory $(\mathbf{x}_T^k, \mathbf{x}_{T-1}^k, \ldots, \mathbf{x}_0^k)$ where $\mathbf{x}_t^k$ is the intermediate state at denoising step $t$.
The per-step importance ratio and clipping operator are defined as:
\begin{equation}
\label{eq:grpo_shorthands}
\rho_t^k(\theta) = \frac{\pi_\theta(\mathbf{x}_{t-1}^k \mid \mathbf{x}_t^k, p)}{\pi_{\theta_{\mathrm{old}}}(\mathbf{x}_{t-1}^k \mid \mathbf{x}_t^k, p)},
\quad
\mathrm{clip}_\varepsilon(\rho) = \mathrm{clip}(\rho,\,1{-}\varepsilon,\,1{+}\varepsilon).
\end{equation}
Each final clean sample $\mathbf{x}_0^k$ is scored by the reward function $r$, and the group-relative advantage is computed as:
\begin{equation}
\label{eq:grpo_advantage}
A^k = \frac{r(\mathbf{x}_0^k,\, p) - \mu_r}{\sigma_r},
\end{equation}
with $\mu_r$ and $\sigma_r$ the mean and standard deviation of $\{r(\mathbf{x}_0^k, p)\}_{k=1}^{K}$.
The policy is then updated by maximizing the clipped surrogate objective:
\begin{equation}
\label{eq:grpo_obj}
\mathcal{J}_{\mathrm{GRPO}}(\theta)
= \frac{1}{K}\sum_{k=1}^{K}\frac{1}{T}\sum_{t=0}^{T-1}
\Bigl[\min\bigl(\rho_t^k(\theta)\,A^k,\;
\mathrm{clip}_\varepsilon(\rho_t^k(\theta))\,A^k\bigr)
- \beta\,D_{\mathrm{KL}}(\pi_{\theta}\,\|\,\pi_{\mathrm{ref}})\Bigr].
\end{equation}
A full description of the framework, including the ODE-to-SDE conversion, closed-form importance ratio and KL divergence, as well as denoising reduction strategy, is provided in~\Cref{app:supp_grpo}.
In~\ourmodel, we instantiate this framework in latent space with geometry-aware rewards computed directly from latents (\Cref{sec:stitching,sec:latentgrpo}).

\noindent\textbf{Geometry Foundation Models} 
are feed-forward transformers~\citep{wang2025vggt} that learn strong geometric priors from large-scale 3D-annotated data with minimal explicit 3D inductive biases.
Given a sequence of $N$ RGB frames $\{\mathbf{I}_i\}_{i=1}^N$ observing the same scene, a geometry model $\mathrm{\Phi}$ predicts per-frame geometric representations
\begin{equation}
\label{eq:geometry_model}
\{\mathbf{O}_i\}_{i=1}^N = \mathrm{\Phi}\big(\{\mathbf{I}_i\}_{i=1}^N\big), 
\qquad 
\mathbf{O}_i = \{\mathbf{C}_i, \mathbf{D}_i, \mathbf{P}_i\}.
\end{equation}
The per-frame outputs $\mathbf{O}_i$ typically include camera pose $\mathbf{C}_i$ (\eg, rotation and translation), a depth map $\mathbf{D}_i$, and a 3D point map $\mathbf{P}_i$ expressed in a shared reference frame.
Although these quantities are geometrically related, jointly predicting them during training has been shown to yield substantial performance gains.
More recent models~\citep{karhade2025any4d, v-dpm, zhu2026motioncrafter, jiang2025geo4d} extend this formulation to dynamic 4D reconstruction by additionally predicting dynamic point maps or scene flow $\mathbf{F}_i$ within $\mathbf{O}_i$, enabling separation of static and dynamic components in the scene.

\subsection{Latent Geometry Model}
\label{sec:stitching}
Geometry foundation models~\citep{wang2025vggt,karhade2025any4d,lin2025depth,keetha2025mapanything} provide strong priors for scene geometry, but they operate in pixel space.
Using them for reward computation therefore requires repeated VAE decoding of diffusion latents, resulting in substantial compute and memory overhead~\citep{mi2025video}.
To eliminate this bottleneck, we construct a \emph{Latent Geometry Model} by stitching video diffusion latents to a pretrained geometry foundation model, enabling geometry prediction directly from latents and allowing rewards to be computed in latent space.

Specifically, let $\mathcal{E}$ denote the encoder of a video VAE~\citep{wan2025wan}, which maps a video $\mathbf{x} = \{\mathbf{I}_i\}_{i=1}^N$ to latents $\mathbf{z} = \mathcal{E}(\mathbf{x})$.
We denote by $\mathrm{\Phi}$ a pretrained geometry model~\citep{wang2025vggt,karhade2025any4d} composed of $L$ transformer layers $\{T_\ell\}_{\ell=1}^L$,
\begin{equation}
\mathrm{\Phi} = T_L \circ T_{L-1} \circ \cdots \circ T_1,
\end{equation}
which maps RGB sequences $\mathbf{x}$ to geometric predictions $\{\mathbf{O}_i\}_{i=1}^N$ as defined in~\Cref{eq:geometry_model}.
For convenience, we define the subnetwork spanning layers $i$ to $j$ as $\mathrm{\Phi}_{i:j} = T_j \circ T_{j-1} \circ \cdots \circ T_i$.

To bypass the RGB input pathway, we replace the first $\hat{\ell}$ layers of $\mathrm{\Phi}$ with a learned 3D convolutional connector $S_\psi$, parameterized by $\psi$, that maps VAE latents directly into the intermediate feature space, giving the latent geometry model:
\begin{equation}
\hat{\mathrm{\Phi}}_\psi = \mathrm{\Phi}_{\hat{\ell}+1:L} \circ S_\psi.
\end{equation}
The stitching layer $\hat{\ell}$ and the parameters $\psi$ are found jointly by minimizing the feature alignment error over a calibration set of $M$ videos $\mathbf{x}^1,\ldots,\mathbf{x}^M$:
\begin{equation}
\label{eq:stitching}
\hat{\ell},\, \psi = \underset{\ell \in \{1,\dots,L\},\; \psi}{\arg\min}\; \frac{1}{M}\sum_{m=1}^{M} \bigl\| S_\psi(\mathcal{E}(\mathbf{x}^m)) - \mathrm{\Phi}_{1:\ell}(\mathbf{x}^m) \bigr\|_2^2.
\end{equation}
We then fine-tune $\psi$ together with the downstream layers $\mathrm{\Phi}_{\hat{\ell}+1:L}$ to further reduce residual discrepancies between $\hat{\mathrm{\Phi}}_\psi$ and the original geometry model $\mathrm{\Phi}$.
Given RGB inputs $\mathbf{x}$, we minimize an alignment loss between their geometric predictions: 
\begin{equation}
\mathcal{L}_{\text{align}}(\psi) 
= \sum_{j} \lambda_j 
\| \hat{\mathrm{\Phi}}_{\psi,j}(\mathcal{E}(\mathbf{x})) - \mathrm{\Phi}_j(\mathbf{x}) \|_1,
\end{equation}
where $j$ indexes predicted geometry modalities (\eg, pose $\mathbf{C}$, depth $\mathbf{D}$, point map $\mathbf{P}$, and scene flow $\mathbf{F}$), and $\lambda_j$ are balancing weights.
Afterwards, the resulting latent geometry model $\hat{\mathrm{\Phi}}_\psi$ produces geometric predictions directly from latent representations:
\begin{equation}
\label{eq:latent_reward_outputs}
\{\mathbf{C}_i, \mathbf{D}_i, \mathbf{P}_i, \mathbf{F}_i\}_{i=1}^{N}
= \hat{\mathrm{\Phi}}_\psi(\mathbf{z}),
\end{equation}
where $\mathbf{F}_i$ is present when $\hat{\mathrm{\Phi}}_\psi$ is constructed from geometry models with dynamic 4D capability~\citep{karhade2025any4d}.
We use these predictions to define geometry-aware rewards in~\Cref{sec:latentgrpo} without RGB decoding, substantially improving efficiency during reinforcement optimization.

\subsection{VGGRPO Training}
\label{sec:latentgrpo}
With the latent geometry model $\hat{\mathrm{\Phi}}_\psi$, we perform \emph{latent-space GRPO} for geometry-aware video post-training, avoiding the compute and memory overhead of repeated VAE decoding in group-based updates. 
In practice, we observe that geometric inconsistency in generated videos is often driven by two factors: (i) jittery or unstable camera motion that induces temporal distortions and structural artifacts, and (ii) inconsistent 3D structure across views, where the same scene content is not geometrically aligned over time. 
Accordingly, our objective is \emph{world-consistent video generation}, which requires both temporally smooth camera trajectories and cross-view coherent scene geometry. 
To this end, we define two complementary rewards from the geometry predicted by $\hat{\mathrm{\Phi}}_\psi$ in~\Cref{eq:latent_reward_outputs}: a \emph{camera motion smoothness} reward and a \emph{geometry reprojection consistency} reward. 
Together, they promote stable camera motion and cross-view geometric coherence, supporting world-consistent generation.

\noindent\textbf{Camera Motion Smoothness Reward.}
To encourage stable and physically plausible camera motion, we define a smoothness reward based on the camera poses $\mathbf{C}_i$ predicted from the denoised video latents $\mathbf{z}_0$ by $\hat{\mathrm{\Phi}}_\psi(\mathbf{z}_0)$.
From $\mathbf{C}_i$ we extract world-frame camera centers $\mathbf{c}_i$ and compute discrete velocities $\mathbf{v}_i = \mathbf{c}_{i+1} - \mathbf{c}_i$ and accelerations $\mathbf{a}_i = \mathbf{v}_i - \mathbf{v}_{i-1}$.
Translational smoothness is then measured by the scale-normalized acceleration:
\begin{equation}
\label{eq:trans_smooth}
e_{\mathrm{trans}}(\mathbf{z}_0)
= \frac{1}{T-2}\sum_{i=2}^{T-1}
\frac{\|\mathbf{a}_i\|_2}
{\|\mathbf{v}_i\|_2+\|\mathbf{v}_{i-1}\|_2}.
\end{equation}
Smooth, near-constant-velocity trajectories yield $e_{\mathrm{trans}}(\mathbf{z}_0) \approx 0$, while jittery motion produces larger values.
Rotational smoothness is measured identically, replacing translational quantities with angular velocities $\boldsymbol{\omega}_i = \log_{SO(3)}(\mathbf{R}_i^\top \mathbf{R}_{i+1})$ and angular accelerations $\boldsymbol{\alpha}_i = \boldsymbol{\omega}_i - \boldsymbol{\omega}_{i-1}$:
\begin{equation}
\label{eq:rot_smooth}
e_{\mathrm{rot}}(\mathbf{z}_0)
= \frac{1}{T-2}\sum_{i=2}^{T-1}
\frac{\|\boldsymbol{\alpha}_i\|_2}
{\|\boldsymbol{\omega}_i\|_2 + \|\boldsymbol{\omega}_{i-1}\|_2}.
\end{equation}
Similarly, $e_{\mathrm{rot}}(\mathbf{z}_0) \approx 0$ for steady rotations and grows with abrupt orientation changes.
The combined motion reward is:
\begin{equation}
\label{eq:motion_reward}
r_{\mathrm{motion}}(\mathbf{z}_0) = \frac{1}{2}\left(\frac{1}{1 + e_{\mathrm{trans}}(\mathbf{z}_0)} + \frac{1}{1 + e_{\mathrm{rot}}(\mathbf{z}_0)}\right).
\end{equation}
Both error terms are mapped to $[0,1]$ via $e\mapsto1/(1+e)$, so $r_{\mathrm{motion}}$ is close to $1$ for smooth trajectories and decreases toward $0$ as jitter increases.

\noindent\textbf{Geometry Reprojection Consistency Reward.}
We quantify cross-view geometric coherence using the point maps $\mathbf{P}_i$, depths $\mathbf{D}_i$, camera parameters $\mathbf{C}_i$, and scene flow $\mathbf{F}_i$ predicted by $\hat{\mathrm{\Phi}}_\psi(\mathbf{z}_0)$, by reprojecting the predicted 3D structure into each view and comparing depths.
We first construct a scene point cloud from the world-frame point maps $\{\mathbf{P}_i\}_{i=1}^N$.
For static scenes, we aggregate all points across frames.
For dynamic scenes, we use the predicted scene flow $\mathbf{F}_i$ to filter dynamic regions and aggregate only static points to obtain a stable scene representation.
We project the resulting point cloud into view $i$ using the predicted camera parameters $\mathbf{C}_i$, producing a rendered depth map $\hat{\mathbf{D}}_i$.
We compare $\hat{\mathbf{D}}_i$ with the predicted depth $\mathbf{D}_i$ over valid projected pixels:
\begin{equation}
\label{eq:depth_reproj_per_view}
e_{\mathrm{geo}}^{(i)}(\mathbf{z}_0)
=
\frac{1}{|\Omega_i|}
\sum_{\mathbf{p}\in\Omega_i}
\left|
\hat{\mathbf{D}}_i(\mathbf{p}) - \mathbf{D}_i(\mathbf{p})
\right|,
\end{equation}
where $\Omega_i$ is the set of valid projected pixels in view $i$.
To focus on local failure cases, we define the geometry reward as the negated average error over the 3 worst views:
\begin{equation}
\label{eq:geometry_reward}
r_{\mathrm{geo}}(\mathbf{z}_0) = -\frac{1}{3}\sum_{i \in \text{top-3}} e_{\mathrm{geo}}^{(i)}(\mathbf{z}_0).
\end{equation}

\noindent\textbf{Alignment Policy Update.} 
For each prompt, we sample $K$ latent videos $\{\mathbf{z}_0^k\}_{k=1}^{K}$ from the current policy and score each with both rewards.
Since $r_{\mathrm{motion}}$ and $r_{\mathrm{geo}}$ have different scales, we normalize each separately within the group and form the advantage as their average, where $\mu_{\mathrm{motion}}, \sigma_{\mathrm{motion}}$ (resp.\ $\mu_{\mathrm{geo}}, \sigma_{\mathrm{geo}}$) denote the mean and standard deviation of each reward across the $K$ samples:
\begin{equation}
\label{eq:combined_adv}
A^k = \frac{1}{2}\left(\frac{r_{\mathrm{motion}}(\mathbf{z}_0^k) - \mu_{\mathrm{motion}}}{\sigma_{\mathrm{motion}}} + \frac{r_{\mathrm{geo}}(\mathbf{z}_0^k) - \mu_{\mathrm{geo}}}{\sigma_{\mathrm{geo}}}\right).
\end{equation}
Substituting $A^k$ into the GRPO objective (\Cref{eq:grpo_obj}) and computing importance ratios $\rho_t^k(\theta)$ over denoised latents $\mathbf{z}_t^k$ (in place of RGB frames $\mathbf{x}_t^k$), we maximize:
\begin{equation}
\label{eq:latent_grpo_obj}
\mathcal{L}_{\mathrm{VGGRPO}}(\theta)
= \frac{1}{K}\sum_{k=1}^{K}\frac{1}{T}\sum_{t=0}^{T-1}
\Bigl[\min\bigl(\rho_t^k(\theta)\, A^k,\;
\mathrm{clip}_\varepsilon(\rho_t^k(\theta))\, A^k\bigr)
- \beta\, D_{\mathrm{KL}}(\pi_{\theta} \| \pi_{\mathrm{ref}})\Bigr].
\end{equation}
We stress that all rewards are computed from $\hat{\mathrm{\Phi}}_\psi(\mathbf{z}_0)$ without decoding RGB frames, yielding an efficient geometry-aware post-training pipeline.

\section{Experiments}

We evaluate the aligned model across diverse scenarios, including static-scene, dynamic-scene, and standard benchmarks, demonstrating its effectiveness and generalization.

\subsection{Experimental Setup}
\noindent\textbf{Implementation Details.}
We construct the latent geometry model by stitching to Any4D~\citep{karhade2025any4d}, a geometry foundation model that supports dynamic 4D reconstruction. 
The latent geometry model is trained for 20 epochs on a mixture of videos generated by the base diffusion model~\citep{wan2025wan} and real videos from DL3DV~\citep{dl3dv}, RealEstate10K~\citep{re10k}, and MiraData~\citep{ju2024miradata}. 
For VGGRPO training, we fine-tune two text-to-video diffusion backbones at different scales, Wan2.1-1B and Wan2.2-5B~\citep{wan2025wan}, using LoRA~\citep{lora} (rank $r=32$, scaling factor $\alpha=64$), with group size $G=64$ and AdamW optimization (learning rate $1\times10^{-4}$, weight decay $1\times10^{-4}$).
Training prompts are sampled from DL3DV, RealEstate10K, and MiraData, following prior geometry-alignment practice~\citep{kupyn2025epipolar}. 
Additional details are provided in~\Cref{sec:supp_training}.

\noindent\textbf{Baselines.}
We compare VGGRPO against representative baselines following prior work~\citep{kupyn2025epipolar,du2026videogpa}. 
For each baseline, we report the best-performing checkpoint to ensure fair comparison. 
We evaluate \textbf{Base Model}, the pretrained generator without post-training; \textbf{Supervised Fine-Tuning (SFT)}, flow-matching fine-tuning on real videos as a purely data-driven adaptation baseline; and two DPO alignment methods: \textbf{Epipolar-DPO}~\citep{kupyn2025epipolar}, which constructs preferences using classical epipolar geometry errors, and \textbf{VideoGPA}~\citep{du2026videogpa}, which derives preferences from reprojection-based consistency scores using VGGT \citep{wang2025vggt}.

\begin{table*}[tb]
  \footnotesize
  \centering
  \renewcommand{\arraystretch}{1.23}%
  \resizebox{\textwidth}{!}{%
  \begin{tabular}{
    p{2.3cm}
      *{5}{>{\centering\arraybackslash}m{0.9cm}}
      *{6}{>{\centering\arraybackslash}m{1.05cm}}
  }
    \toprule
    \multirow{2}{*}{\textbf{Method}}
      & \multicolumn{3}{c}{\textbf{Static}}
      & \multicolumn{2}{c}{\textbf{Dynamic}}
      & \multirow{2}{*}[-0.4ex]{\parbox{1.\linewidth}{\centering \textbf{Sub.\\Cons.}\,$\uparrow$}}
      & \multirow{2}{*}[-0.4ex]{\parbox{1.05\linewidth}{\centering \textbf{Bg.\\Cons.}\,$\uparrow$}}
      & \multirow{2}{*}[-0.4ex]{\parbox{1.05\linewidth}{\centering \textbf{Aes.\\Qual.}\,$\uparrow$}} 
      & \multirow{2}{*}[-0.4ex]{\parbox{1.05\linewidth}{\centering \textbf{Img.\\Qual.}\,$\uparrow$}}
      & \multirow{2}{*}[-0.4ex]{\parbox{1.0\linewidth}{\centering \textbf{Mot.\\Smooth.}\,$\uparrow$}}
      & \multirow{2}{*}[-0.4ex]{\parbox{1.\linewidth}{\centering \textbf{Dyn.\\Deg.}\,$\uparrow$}} 
      \\
        \noalign{\vskip-0.4ex}
        \cmidrule(r{0.2em}){2-4}\cmidrule(l{0.2em}){5-6}
        \noalign{\vskip-0.8ex}
        & \rule{0pt}{1.25em}\textbf{VQ}\,$\uparrow$
        & \rule{0pt}{1.25em}\textbf{MQ}\,$\uparrow$
        & \rule{0pt}{1.25em}\textbf{Epi.}\,$\downarrow$
        & \rule{0pt}{1.25em}\textbf{VQ}\,$\uparrow$
        & \rule{0pt}{1.25em}\textbf{MQ}\,$\uparrow$
        &  
        &  
        &  
        &  
        &  
        &  
        \\
    \midrule
    \noalign{\vspace{-0.7ex}}
    \rowcolor{gray!10}
    \multicolumn{12}{l}{\textit{Base Model: Wan2.1-1B}} \\

    Base         &  -      &  -      &  0.133 &  -     &  -     & 0.7941  & 0.8930  &  0.5233 & 0.6178 & 0.9552 & \textbf{0.9231} \\
    SFT          &  45.26  &  46.84  &  0.137 &  40.00  &  39.00  & 0.8032  & 0.8896  &  0.5472 & 0.6256 & 0.9646 & 0.8795 \\
    Epipolar-DPO &  \underline{54.21}  &  55.79  & \textbf{0.098}  &  \underline{45.50}  &  \underline{43.00}  & \underline{0.8125}  & 0.8916  &  \underline{0.5578} & 0.6461 & \underline{0.9671} & 0.8816 \\
    VideoGPA     &  53.68  &  \underline{56.32}  & 0.105  &  42.50  &  41.00  & 0.8068  & \underline{0.8931}  &  0.5562 & \underline{0.6507} & 0.9650 & 0.8734 \\
    \rowcolor{row} 
    VGGRPO~(Ours)&  \textbf{59.47}  &  \textbf{66.84}  & \underline{0.102}  &  \textbf{57.00}  &  \textbf{63.00}  & \textbf{0.8255}  & \textbf{0.8974}  &  \textbf{0.5623} & \textbf{0.6585} & \textbf{0.9753} & \underline{0.9048} \\

    \midrule
    \noalign{\vspace{-0.7ex}}
    \rowcolor{gray!10}
    \multicolumn{12}{l}{\textit{Base Model: Wan2.2-5B}} \\

    Base         &  -      &  -      & 0.142 &  -     &  -     & 0.8151  & 0.8958  &  0.4837 & \underline{0.6402} & 0.9467 & \underline{0.8692} \\
    SFT          &  46.32  &  52.63  & 0.129 &  33.00  &  51.00  & 0.8323  & 0.8925  &  0.4886 & 0.6159 & \underline{0.9548} & \textbf{0.9026} \\
    Epipolar-DPO &  52.11  &  58.95  & 0.101&  38.00  &  \underline{54.50}  & 0.8407  & \underline{0.9054}  &  \underline{0.4945} & 0.6275 & 0.9482 & 0.7603 \\
    VideoGPA     &  \underline{54.74}  &  \underline{60.53}  & \underline{0.098} &  \underline{40.00}  & 54.00   & \underline{0.8511}  & 0.9048  &  0.4920 & 0.6131 & 0.9518 & 0.7645 \\
    \rowcolor{row} 
    VGGRPO~(Ours)&  \textbf{62.63}  &  \textbf{68.42}  &  \textbf{0.093} &  \textbf{56.50}  &  \textbf{66.00}  & \textbf{0.8672}  & \textbf{0.9056}  &  \textbf{0.5094} & \textbf{0.6843} & \textbf{0.9619} & 0.8421 \\

    \bottomrule
  \end{tabular}%
  }
  \vspace{1.0ex}
  \caption{
\textbf{Quantitative Comparison.} 
The best and runner-up results are shown in \textbf{bold} and \underline{underlined}, respectively.
In both static- and dynamic-scene benchmarks, our method consistently improves geometric consistency and overall video quality over baselines across different base models, demonstrating the effectiveness of latent-space geometry-aware post-training.
  }
  \label{tab:main_tab}
\end{table*}

\begin{figure*}[t] 
    \centering
    \includegraphics[width=\textwidth]{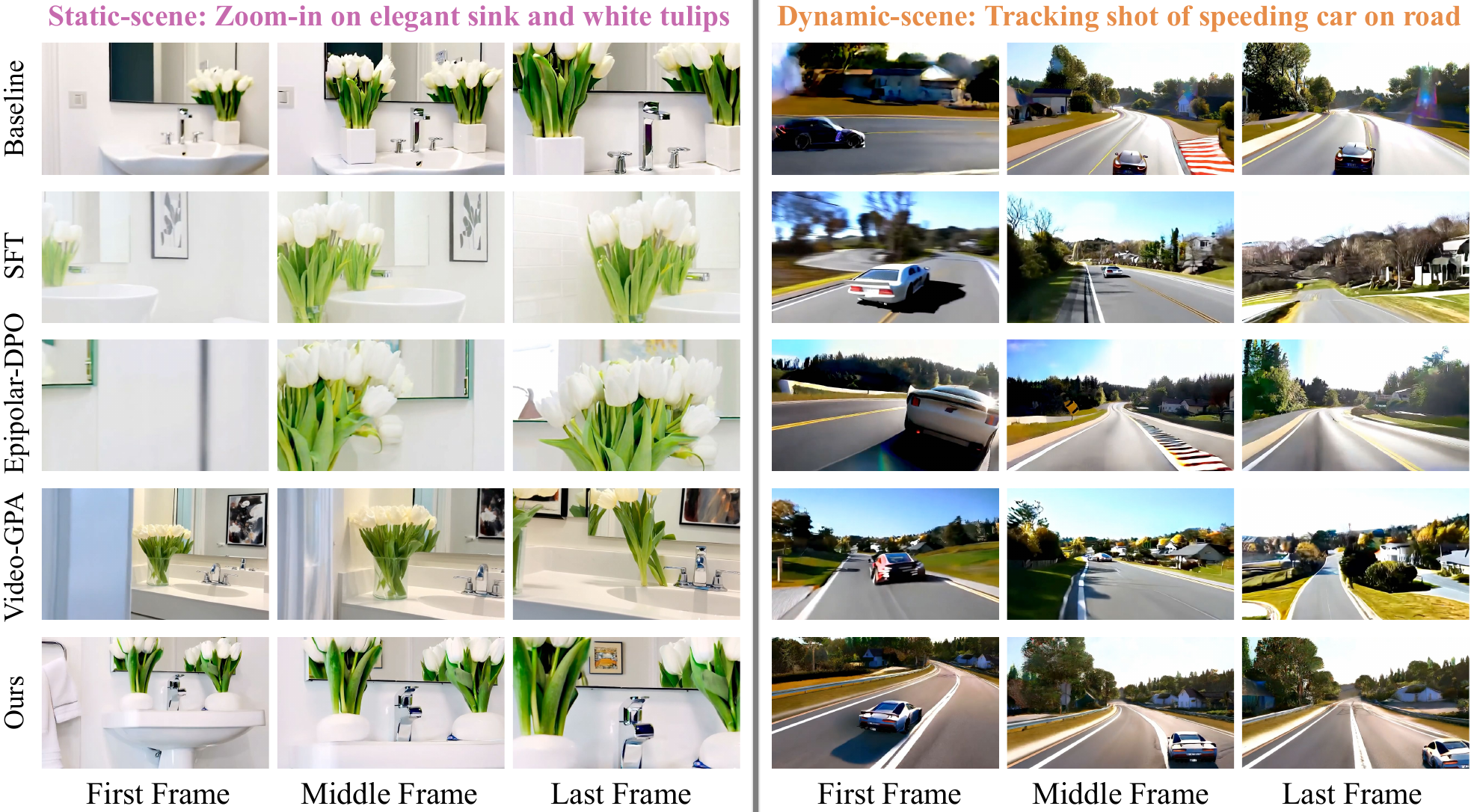} 
    \caption{\textbf{Qualitative Comparison on Static and Dynamic Scenes.} 
    We show the first, middle, and last frames of video generations for a static-scene prompt (\textit{left}) and a dynamic-scene prompt (\textit{right}), with a representative segment of each prompt shown at the top. 
    All baselines exhibit inconsistent artifacts, including geometric drift, temporal flicker, and unstable camera motion. 
    In contrast, VGGRPO produces more coherent scene structure with smoother camera trajectories across frames.
    }
    \label{fig:comparison} 
\end{figure*}

\subsection{Main Results}
\noindent\textbf{Quantitative Evaluation.}
To evaluate post-training across diverse scenarios, we construct two held-out benchmarks: 190 static-scene captions from DL3DV~\citep{dl3dv} and RealEstate10K~\citep{re10k}, and 200 dynamic-scene captions from MiraData~\citep{ju2024miradata}. 
The dynamic split is curated to include challenging cases with significant non-rigid motion, making geometric drift and camera instability more pronounced.
We evaluate from two perspectives: \emph{general video quality} and \emph{geometry-related quality}. 
For general video quality, we follow VBench~\citep{huang2024vbench} and report \textit{subject consistency}, \textit{background consistency}, \textit{motion smoothness}, \textit{aesthetic quality}, \textit{imaging quality}, and \textit{dynamic degree}. 
For geometry-related quality, following prior studies~\citep{kupyn2025epipolar,du2026videogpa}, we report (i) \textit{VideoReward}~\citep{liuimproving} win rates against the base model for \textit{Visual Quality} (VQ) and \textit{Motion Quality} (MQ), which reflect human preference, and (ii) \textit{Sampson epipolar error} on the static split only (where epipolar assumptions hold), measuring point-to-epipolar-line distances as a classical geometric-precision metric.
We report geometry-related metrics separately on the static and dynamic splits, and compute general video quality metrics over all 390 captions.

As shown in~\Cref{tab:main_tab}, VGGRPO consistently outperforms all baselines on geometry-related metrics. 
Across both static and dynamic splits, VGGRPO achieves higher motion quality and stronger geometric consistency, demonstrating the effectiveness of latent-space geometry-aware post-training. 
While prior geometry-aligned baselines~\citep{kupyn2025epipolar,du2026videogpa} can generalize to limited dynamic settings, their performance degrades noticeably on our dynamic benchmark with complex non-rigid motion, where maintaining coherent geometry is substantially harder. 
In contrast, VGGRPO remains robust under significant scene dynamics due to our 4D-aware latent geometry model and reward design. 
VGGRPO also improves VBench metrics overall, indicating that our improved geometric consistency does not come at the expense of perceptual quality.

\noindent\textbf{Qualitative Results.}
\Cref{fig:comparison} presents qualitative comparisons on both static- and dynamic-scene settings, visualized using the first, middle, and last frames of each generated video.
All baselines struggle to maintain world consistency, exhibiting unstable camera motion, geometric drift, and temporally inconsistent scene structure.
In the static-scene example, the base model introduces spurious content (a second flower in the middle frame), while SFT shows signs of overfitting to biases in the limited real data, yielding less faithful structure.
Epipolar-DPO and VideoGPA show noticeable temporal flicker in the early and late frames, respectively, indicating geometric instability.
On the dynamic example, the gap widens: baselines degrade further in both visual quality and geometry, producing less stable frames and inconsistent structure, reflecting reduced robustness under complex dynamic scenes.
In contrast, \ourmodel~yields more coherent scene structure with smoother camera motion in both settings, highlighting the effectiveness of our geometry-guided latent-space post-training.

\begin{table*}[tb]
  \footnotesize
  \centering
  \renewcommand{\arraystretch}{1}
  \begin{subtable}[t]{0.29\textwidth}
    \captionsetup{justification=centering,singlelinecheck=true}
    \centering
    \resizebox{1\linewidth}{!}{%
        \setlength{\tabcolsep}{3pt}
        \renewcommand{\arraystretch}{1.12}
        \begin{tabular}{cccc}
          \toprule
          \textbf{Geo-FM} & \textbf{VQ}\,$\uparrow$ & \textbf{MQ}\,$\uparrow$ & \textbf{Epi.}\,$\downarrow$ \\
          \midrule
            VGGT                 &  54.96 & 60.61  & \textbf{0.090}  \\
           \rowcolor{row} Any4D  & \textbf{59.57} & \textbf{67.21} & 0.093  \\
          \bottomrule
        \end{tabular}
    }
    \subcaption{Impact of geometry FM.}
    \vspace{0.08in}
    \label{tab:ab_gfm}
  \end{subtable}
  \hfill
  \begin{subtable}[t]{0.31\textwidth}
    \captionsetup{justification=centering,singlelinecheck=true}
    \centering
    \resizebox{1\linewidth}{!}{%
        \setlength{\tabcolsep}{3pt}
        \renewcommand{\arraystretch}{1.15}
        \begin{tabular}{ccccc}
          \toprule
          $r_{\mathrm{motion}}$ & $r_{\mathrm{geo}}$ &\textbf{VQ}\,$\uparrow$ & \textbf{MQ}\,$\uparrow$ & \textbf{Epi.}\,$\downarrow$ \\
          \midrule
               \checkmark    &              &  55.60 & 63.40  &  0.104 \\
           \rowcolor{row} \checkmark & \checkmark & \textbf{59.57} & \textbf{67.21} & \textbf{0.093}  \\
          \bottomrule
        \end{tabular}
    }
    \subcaption{Impact of reward terms.}
    \vspace{0.08in}
    \label{tab:ab_rewards}
  \end{subtable}
  \hfill
  \begin{subtable}[t]{0.39\textwidth}
    \captionsetup{justification=centering,singlelinecheck=true}
    \centering
    \resizebox{1\linewidth}{!}{%
        \setlength{\tabcolsep}{3pt}
        \renewcommand{\arraystretch}{1.12}
        \begin{tabular}{ccccc}
          \toprule
                  & \textbf{VQ}\,$\uparrow$ & \textbf{MQ}\,$\uparrow$ & \textbf{Epi.}\,$\downarrow$ & \textbf{Time}\,$\downarrow$ \\
          \midrule
            Baseline                   & - & - & 0.142 & \textbf{44.35} \\
           \rowcolor{row} + guidance  & \textbf{52.63} & \textbf{52.37} & \textbf{0.136} & 62.60\\
          \bottomrule
        \end{tabular}
    }
    \subcaption{Test-time reward guidance.}
    \vspace{0.08in}
    \label{tab:ab_guidance}
  \end{subtable}
  \begin{subtable}[t]{0.76\textwidth}
    \captionsetup{justification=centering,singlelinecheck=true}
    \centering
    \resizebox{1\linewidth}{!}{%
        \setlength{\tabcolsep}{3pt}
        \renewcommand{\arraystretch}{1.32}
        \begin{tabular}{ccccccc}
          \toprule
          
      \textbf{Model}
      & \textbf{Sub. Cons.}\,$\uparrow$
      & \textbf{Bg. Cons.}\,$\uparrow$
      & \textbf{Aes. Qual.}\,$\uparrow$ 
      & \textbf{Img. Qual.}\,$\uparrow$
      & \textbf{Mot. Smooth.}\,$\uparrow$
      & \textbf{Dyn. Deg.}\,$\uparrow$  \\
          \midrule
           Baseline              &  0.9542 & 0.9528  & 0.5966 & 0.6733  & 0.9841 & \textbf{0.4237}  \\
           \rowcolor{row} Ours   &  \textbf{0.9644} & \textbf{0.9583}  & \textbf{0.5991} & \textbf{0.6861}  & \textbf{0.9895} & 0.3962  \\
          \bottomrule
        \end{tabular}
    }
    \subcaption{Generalization performance on standard VBench captions.}
    \label{tab:ab_vbench}
  \end{subtable}
  \begin{subtable}[t]{0.23\textwidth}
    \captionsetup{justification=centering,singlelinecheck=true}
    \centering
    \resizebox{1\linewidth}{!}{%
        \setlength{\tabcolsep}{3pt}
        \renewcommand{\arraystretch}{1.4}
        \begin{tabular}{ccc}
          \toprule
             \textbf{Reward}         & \textbf{Time}\,$\downarrow$ & \textbf{Mem}\,$\downarrow$  \\
          \midrule
            RGB-based                 & 54.73 & 76.80 \\
            \rowcolor{row} Ours        & \textbf{41.33} & \textbf{68.57}  \\
          \bottomrule
        \end{tabular}
    }
    \subcaption{Efficiency study.}
    \label{tab:eff}
  \end{subtable}
\caption{\textbf{Additional Studies.}
(a) \textbf{Impact of geometry foundation models:} VGGRPO generalizes across latent geometry models constructed from different geometry foundation models.
(b) \textbf{Impact of reward components:} $r_{\mathrm{motion}}$ and $r_{\mathrm{geo}}$ provide complementary gains, and their combination yields the best overall performance.
(c) \textbf{Test-time reward guidance:} Differentiable latent-space guidance through the latent geometry model improves geometric consistency at test time with modest runtime overhead and no training.
(d) \textbf{Generalization:} Beyond improving world consistency, VGGRPO transfers to standard VBench captions, indicating improved generation quality and demonstrating robustness in diverse scenarios.
(e) \textbf{Efficiency study:} Latent-space reward computation reduces both runtime and peak GPU memory compared to RGB-based rewarding.
VQ/MQ denote \textit{VideoReward} win rates for Visual and Motion Quality, Epi.\ denotes the Sampson epipolar error on static scenes.
Time is measured in seconds, and Mem denotes peak GPU memory (GB).
Best results are in \textbf{bold}.}
\end{table*}

\begin{figure*}[t] 
    \centering
    \includegraphics[width=\textwidth]{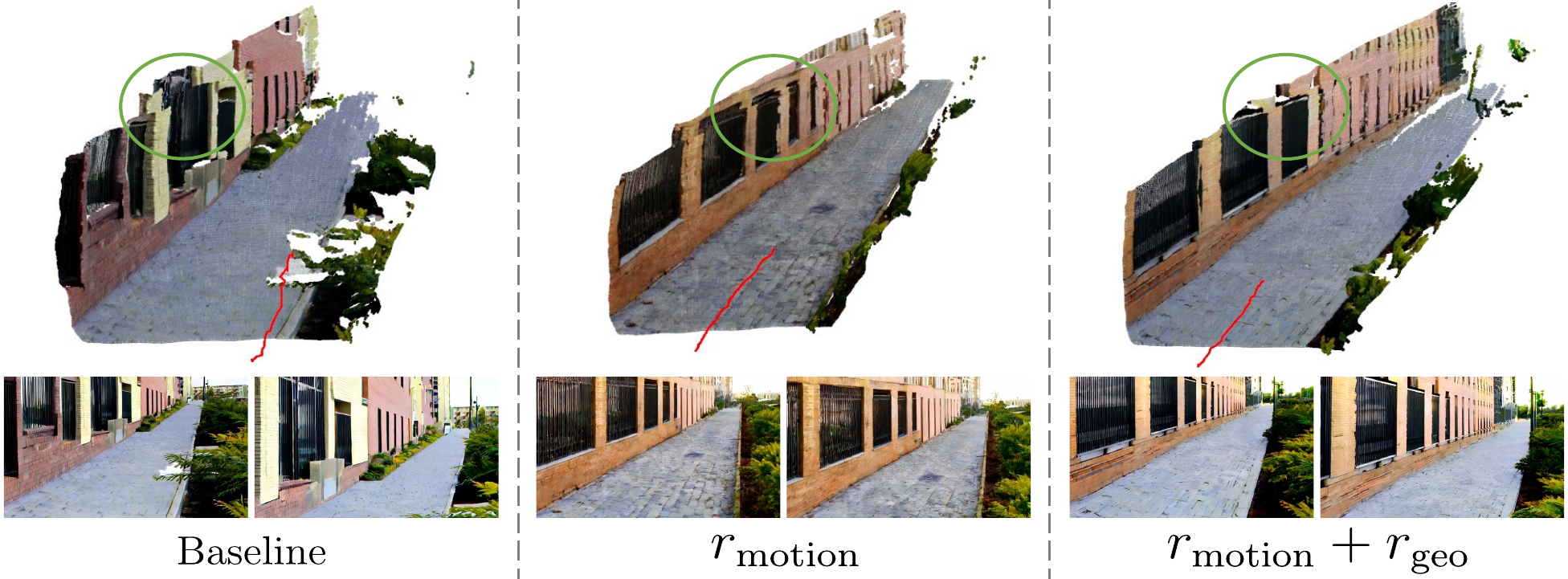} 
\caption{\textbf{Reward Components Ablation.}
The reconstructed scene visualizes the estimated camera trajectory (red curve), with the first and last frames shown below each reconstruction.
Compared to the Baseline, optimizing the motion reward $r_{\mathrm{motion}}$ stabilizes camera motion, but geometric artifacts persist (green circle).
Adding the reprojection consistency reward ($r_{\mathrm{motion}}{+}r_{\mathrm{geo}}$) further improves scene geometry while preserving smooth camera motion, proving the complementarity of both components.
}
    \vspace{-0.8ex} 
    \label{fig:ab_motion} 
\end{figure*}

\subsection{Additional Studies}
\label{sec:ablation}
Unless otherwise stated, we use Wan2.2-5B~\citep{wan2025wan} as the base model and report \textit{VideoReward} win rates for Visual Quality and Motion Quality, averaged over captions from both the static- and dynamic-scene splits, as well as the Sampson epipolar error (Epi.) on the static split to quantify geometric alignment.

\noindent\textbf{Impact of geometry foundation models.}
\Cref{tab:ab_gfm} evaluates the effect of using different geometry foundation models~\citep{wang2025vggt,karhade2025any4d} to construct the latent geometry model. 
When stitching to VGGT~\citep{wang2025vggt}, VGGRPO training uses only static-scene captions to respect VGGT's static-scene assumptions. 
As shown in~\Cref{tab:ab_gfm}, the VGGT-based variant achieves slightly lower epipolar error on static scenes, while the Any4D-based variant attains higher VQ and MQ by benefiting from its dynamic-scene support.
Both variants outperform prior post-training baselines~\citep{kupyn2025epipolar,du2026videogpa}, indicating that VGGRPO is effective across different geometry foundation models.

\noindent\textbf{Impact of reward components.}
As shown in \Cref{fig:ab_motion}, optimizing only $r_{\mathrm{motion}}$ substantially stabilizes the camera trajectory (red curve) compared to the shaky baseline, while geometric artifacts remain, \eg, the inconsistent wall structure highlighted in the green circle.
Adding $r_{\mathrm{geo}}$ effectively mitigates these reprojection inconsistencies while maintaining smooth camera motion, yielding more coherent scene geometry and improved perceptual quality.
The quantitative results in \Cref{tab:ab_rewards} mirror these observations: $r_{\mathrm{motion}}$ alone improves alignment, and combining it with $r_{\mathrm{geo}}$ achieves the best results, confirming both the necessity and complementarity of the two rewards.

\noindent\textbf{Test-time reward guidance.}
Our latent geometry model is differentiable, enabling gradient-based test-time guidance directly in latent space that steers the model toward improved geometric consistency without RGB decoding.
See~\Cref{sec:supp_training} for more implementation details.
\Cref{tab:ab_guidance} shows that applying reward guidance once every 20 denoising steps (out of 50 total) consistently improves geometric consistency with modest runtime overhead and no training, providing a training-free way to enhance geometry for a fixed video diffusion model.
While the gains are smaller than those from full VGGRPO post-training, these results highlight the utility of our latent geometry model and reward formulation beyond training-time alignment.

\noindent\textbf{Generalization.}
Beyond geometric consistency, we also verify that VGGRPO preserves the quality of general-purpose video generation.
As shown in \Cref{tab:ab_vbench}, when evaluated on the standard VBench caption set~\citep{huang2024vbench}, our model outperforms the baseline across most metrics, while slightly lower on Dynamic Degree.
We attribute this to the definition of Dynamic Degree, which is computed based on the magnitude of RAFT optical flow~\citep{teed2020raft}: by explicitly encouraging smoother camera trajectories, our motion reward reduces camera jitter and thus lowers the flow magnitude, even as perceptual motion quality improves.
Importantly, these gains are achieved without specific training targeting general video quality, indicating that our geometry-aligned post-training preserves the model’s original generalization.
Overall, VGGRPO acts as a broadly applicable regularizer that improves perceptual quality and remains robust across diverse scenarios.

\noindent\textbf{Efficiency study.}
\Cref{tab:eff} reports the wall-clock time (seconds) and peak GPU memory (GB) of the reward computation step (batch size 4) for RGB-based rewarding and our latent-based rewarding.
Compared to RGB-based, our latent reward reduces peak memory from 76.80\,GB to 68.57\,GB and speeds up reward computation from 54.73\,s to 41.33\,s (13.40\,s faster, 24.5\% reduction), demonstrating the efficiency of computing geometry rewards directly in latent space.

\section{Conclusion}
We introduce VGGRPO, a geometry-aware post-training framework that aligns pretrained video diffusion models toward 4D world-consistent generation by optimizing directly in latent space.
To this end, we develop a latent geometry model that connects video latents with geometry foundation models, leveraging their strong geometric priors to predict 4D scene geometry for reward computation.
Building on this model, we perform latent-space GRPO, enabling efficient group-based policy updates without repeated VAE decoding.
We further design two complementary rewards, camera motion smoothness and geometry reprojection consistency, which jointly promote temporally stable camera trajectories and cross-view coherent scene structure. 
This formulation applies to both static- and dynamic-scene videos, encouraging 4D world-consistent video generation.
Across both static and dynamic benchmarks, \ourmodel\ substantially improves geometric consistency and camera smoothness while preserving strong visual fidelity, 
demonstrating that latent-space 4D geometry rewards provide an efficient and scalable approach for aligning video generation models.

\section*{Acknowledgements}
We would like to thank Hyojun Go (ETH Zurich), Dominik Narnhofer (ETH Zurich), Nikolai Kalischek (Google), Mattia Segu (Google), and Fabian Manhardt (Google) for their valuable support and thoughtful discussion during the project. 
Zhaochong An and Serge Belongie are supported by funding from the Pioneer Centre for AI, DNRF grant number P1. Orest Kupyn is supported by a Google unrestricted gift.

\bibliographystyle{plainnat}
\nobibliography*
\bibliography{example_paper}

\newpage
\appendix
\crefalias{section}{appendix}

\onecolumn

\section{Flow-Based Group Relative Policy Optimization}
\label{app:supp_grpo}

In this section we provide a self-contained account of the reinforcement learning framework that underpins~\ourmodel.
We first recall the rectified flow formulation used by modern video diffusion models (\Cref{sec:supp_rectflow}), then show how the iterative denoising process can be cast as a Markov decision process (\Cref{sec:supp_mdp}), describe how Group Relative Policy Optimization is adapted to flow matching models (\Cref{sec:supp_grpo_detail}), detail the ODE-to-SDE conversion that enables stochastic exploration (\Cref{sec:supp_ode2sde}), and finally discuss the closed-form KL divergence and the denoising reduction strategy that make training practical (\Cref{sec:supp_kl,sec:supp_denred}).

\subsection{Rectified Flow}
\label{sec:supp_rectflow}

Let $\mathbf{x}_0$ denote a clean data sample (\eg, a video encoded as a latent tensor) and $p$ the associated text prompt.
Rectified flow~\citep{rectified_flow} constructs a transport path between the data distribution and a standard Gaussian by defining the forward interpolation
\begin{equation}
\label{eq:supp_forward}
\mathbf{x}_t = (1 - t)\,\mathbf{x}_0 + t\,\epsilon, \qquad t \in [0, 1],\quad \epsilon \sim \mathcal{N}(\mathbf{0}, \mathbf{I}).
\end{equation}
At $t=0$ we recover the clean sample $\mathbf{x}_0$, and at $t=1$ we obtain pure noise.
A neural network parameterizes a velocity field $v_\theta(\mathbf{x}_t, t, p)$ that is trained to regress the ground-truth transport direction $\mathbf{v} = \epsilon - \mathbf{x}_0$ by minimizing the conditional flow matching objective~\citep{rectified_flow}:
\begin{equation}
\label{eq:supp_fm_loss}
\mathcal{L}_{\mathrm{FM}}(\theta) = \mathbb{E}_{t \sim \mathcal{U}[0,1],\;\mathbf{x}_0,\;\epsilon}\bigl[\|(\epsilon - \mathbf{x}_0) - v_\theta(\mathbf{x}_t, t, p)\|^2\bigr].
\end{equation}
Once trained, sampling proceeds by integrating the learned ODE
\begin{equation}
\label{eq:supp_ode}
\frac{\mathrm{d}\mathbf{x}_t}{\mathrm{d}t} = v_\theta(\mathbf{x}_t, t, p)
\end{equation}
backward from $t=1$ (noise) to $t=0$ (data), using a numerical solver such as Euler's method with a discrete set of $T$ timesteps $1 = t_T > t_{T-1} > \cdots > t_0 = 0$.
The rectified flow formulation encourages nearly straight-line trajectories between noise and data, enabling high-quality generation with relatively few solver steps and serving as the backbone of state-of-the-art video diffusion models~\citep{wan2025wan,hunyanvideo}.

\subsection{Denoising as a Multi-Step MDP}
\label{sec:supp_mdp}

Most use cases of generative models fine-tuning are not directly concerned with matching a data distribution, but rather with downstream objectives such as geometric consistency or aesthetic quality.
To optimize such objectives, DDPO~\citep{ddpo} recasts the iterative denoising process as a multi-step Markov Decision Process (MDP) $(\mathcal{S}, \mathcal{A}, \rho_0, P, R)$, where:
\begin{equation}
\label{eq:supp_mdp}
\begin{aligned}
\mathbf{s}_t &\triangleq (p,\, t,\, \mathbf{x}_t), &
\mathbf{a}_t &\triangleq \mathbf{x}_{t-1}, \\
\pi(\mathbf{a}_t \mid \mathbf{s}_t) &\triangleq p_\theta(\mathbf{x}_{t-1} \mid \mathbf{x}_t, p), &
\rho_0(\mathbf{s}_0) &\triangleq \bigl(p(p),\, \delta_T,\, \mathcal{N}(\mathbf{0}, \mathbf{I})\bigr), \\
P(\mathbf{s}_{t+1} \mid \mathbf{s}_t, \mathbf{a}_t) &\triangleq (\delta_p,\, \delta_{t-1},\, \delta_{\mathbf{x}_{t-1}}), &
R(\mathbf{s}_t, \mathbf{a}_t) &\triangleq
\begin{cases} r(\mathbf{x}_0, p) & \text{if } t = 0, \\ 0 & \text{otherwise.}\end{cases}
\end{aligned}
\end{equation}
Here $\delta_{(\cdot)}$ denotes a Dirac delta, the state $\mathbf{s}_t$ comprises the prompt, current timestep, and noisy sample, the action $\mathbf{a}_t$ is the denoised prediction at the next step, and the transition $P$ is deterministic (the context and timestep advance deterministically once the action is taken).
The reward $r(\mathbf{x}_0, p)$ is an arbitrary downstream reward function evaluated only on the final clean sample.

The key insight of this formulation is that, unlike the intractable marginal likelihood $\log p_\theta(\mathbf{x}_0 \mid p)$ of the full chain, each individual denoising step $p_\theta(\mathbf{x}_{t-1} \mid \mathbf{x}_t, p)$ has an exact, tractable log-probability when the reverse kernel is Gaussian (see~\Cref{sec:supp_ode2sde}).
This means standard policy gradient estimators can be applied at every step of the denoising trajectory, yielding unbiased gradient estimates for the RL objective:
\begin{equation}
\label{eq:supp_rl_obj}
\mathcal{J}_{\mathrm{DDRL}}(\theta) = \mathbb{E}_{p \sim \mathcal{P},\;\mathbf{x}_0 \sim \pi_\theta(\cdot \mid p)}\bigl[r(\mathbf{x}_0, p)\bigr],
\end{equation}
where $\mathcal{P}$ denotes the distribution over text prompts.
This objective enables optimizing diffusion and flow models for any black-box reward without requiring the reward to be differentiable.

\subsection{Group Relative Policy Optimization}
\label{sec:supp_grpo_detail}

While standard policy gradient methods like Proximal Policy Optimization (PPO)~\citep{schulman2017proximal} require learning a separate value function (critic) to estimate per-step advantages, GRPO~\citep{grpo} provides a lightweight alternative that eliminates the critic entirely.
This is particularly beneficial for diffusion and flow models, where the high dimensionality of the state space (each $\mathbf{x}_t$ is a full image or video latent) would make training a value network expensive and unstable.

\paragraph{Group-relative advantage estimation.}
For each prompt $p$, the current policy $\pi_\theta$ generates a group of $K$ samples $\{\mathbf{x}_0^k\}_{k=1}^{K}$ together with their full denoising trajectories $\{(\mathbf{x}_T^k, \mathbf{x}_{T-1}^k, \ldots, \mathbf{x}_0^k)\}_{k=1}^{K}$.
A reward function $r$ scores each final sample, and the advantage of the $g$-th sample is estimated by normalizing the rewards within the group:
\begin{equation}
\label{eq:supp_advantage}
A^k = \frac{r(\mathbf{x}_0^k, p) - \mu_r}{\sigma_r},
\end{equation}
where $\mu_r = \frac{1}{K}\sum_{k=1}^{K} r(\mathbf{x}_0^k, p)$ and $\sigma_r = \mathrm{std}(\{r(\mathbf{x}_0^k, p)\}_{k=1}^{K})$ are the mean and standard deviation of the rewards within the group.
This normalization serves the same role as a learned baseline: it centers the advantages around zero and scales them to unit variance, ensuring that roughly half the samples in each group receive positive updates and half receive negative updates, regardless of the absolute reward scale.
Crucially, $A^k$ is constant across all denoising steps $t$ for a given sample, since the reward is only observed at $t=0$.

\paragraph{Clipped surrogate objective.}
The policy is updated by maximizing a clipped surrogate objective that prevents excessively large policy updates, analogous to PPO's clipping mechanism.
Using the per-step importance ratio $\rho_t^k(\theta)$ and the clipping operator $\mathrm{clip}_\varepsilon$ defined in~\Cref{eq:grpo_shorthands} of the main text, the GRPO objective is (cf.~\Cref{eq:grpo_obj}):
\begin{equation}
\label{eq:supp_grpo_obj}
\mathcal{J}_{\mathrm{GRPO}}(\theta)
= \frac{1}{K}\sum_{k=1}^{K}\frac{1}{T}\sum_{t=0}^{T-1}
\Bigl[\min\bigl(\rho_t^k(\theta)\,A^k,\;
\mathrm{clip}_\varepsilon(\rho_t^k(\theta))\,A^k\bigr)
- \beta\,D_{\mathrm{KL}}(\pi_{\theta}\,\|\,\pi_{\mathrm{ref}})\Bigr].
\end{equation}
The $\min$ operator ensures that, when $A^k > 0$ (the sample is better than the group average), the policy ratio $\rho_t^k(\theta)$ is clipped from above at $1 + \varepsilon$, limiting how much the policy can increase the probability of this trajectory.
Conversely, when $A^k < 0$, the ratio is clipped from below at $1 - \varepsilon$, preventing the policy from decreasing the probability too aggressively.
The KL penalty $\beta\,D_{\mathrm{KL}}(\pi_\theta \| \pi_{\mathrm{ref}})$ regularizes the updated policy toward the reference (pretrained) policy $\pi_{\mathrm{ref}}$, preventing reward hacking and preserving the general capabilities of the base model.

\subsection{ODE-to-SDE Conversion}
\label{sec:supp_ode2sde}

GRPO relies on stochastic sampling to generate diverse trajectories for exploration and advantage estimation.
However, rectified flow models generate samples via a deterministic ODE (\Cref{eq:supp_ode}), which produces a one-to-one mapping between the initial noise and the final sample.
This determinism is problematic in two ways:
(1) the importance ratio $\rho_t^k(\theta)$ requires evaluating the conditional $p_\theta(\mathbf{x}_{t-1} \mid \mathbf{x}_t, p)$, which has no closed form under deterministic dynamics;
(2) the lack of randomness beyond the initial seed severely limits exploration, reducing the diversity of the generated group and the quality of the advantage estimates.

Flow-GRPO~\citep{flowgrpo} resolves this by converting the deterministic ODE into an equivalent SDE that preserves the marginal distribution $p_t(\mathbf{x}_t)$ at every timestep while injecting controlled noise.
The key theoretical result, following~\citet{song2021scorebased}, is that for any ODE with velocity field $v_t$, there exists a family of SDEs indexed by a diffusion coefficient $\sigma_t$ that share the same marginal distributions.
For rectified flow, this SDE takes the form:
\begin{equation}
\label{eq:supp_sde}
\mathrm{d}\mathbf{x}_t
= \underbrace{\biggl[v_t(\mathbf{x}_t) + \frac{\sigma_t^2}{2t}\bigl(\mathbf{x}_t + (1 - t)\,v_t(\mathbf{x}_t)\bigr)\biggr]}_{\text{drift (adjusted)}}\mathrm{d}t
+ \underbrace{\sigma_t\,\mathrm{d}\mathbf{w}}_{\text{diffusion}},
\end{equation}
where $\mathrm{d}\mathbf{w}$ denotes standard Wiener process increments.
The drift term is modified from the original ODE velocity $v_t$ by an additional correction that depends on the score function $\nabla \log p_t(\mathbf{x}_t)$, which for rectified flow can be expressed analytically in terms of $v_t$ and $\mathbf{x}_t$.
The diffusion coefficient $\sigma_t$ controls the amount of stochasticity: setting $\sigma_t = 0$ recovers the original deterministic ODE, while larger values increase exploration at the cost of noisier trajectories.
Following~\citet{flowgrpo}, we use $\sigma_t = a\sqrt{t / (1 - t)}$, where $a$ is a scalar hyperparameter.

Applying Euler--Maruyama discretization to~\Cref{eq:supp_sde} yields the update rule:
\begin{equation}
\label{eq:supp_sde_step}
\mathbf{x}_{t+\Delta t}
= \mathbf{x}_t
+ \biggl[v_\theta(\mathbf{x}_t, t, p) + \frac{\sigma_t^2}{2t}\bigl(\mathbf{x}_t + (1 - t)\,v_\theta(\mathbf{x}_t, t, p)\bigr)\biggr]\Delta t
+ \sigma_t\sqrt{\Delta t}\;\epsilon, \quad \epsilon \sim \mathcal{N}(\mathbf{0}, \mathbf{I}).
\end{equation}
This makes each denoising step a Gaussian transition with mean $\bar{\mathbf{x}}_{t+\Delta t,\theta}$ (the deterministic part) and variance $\sigma_t^2 \Delta t\, \mathbf{I}$, so the reverse-step distribution is
\begin{equation}
\label{eq:supp_reverse_gaussian}
\pi_\theta(\mathbf{x}_{t-1} \mid \mathbf{x}_t, p) = \mathcal{N}\bigl(\mathbf{x}_{t-1};\; \bar{\mathbf{x}}_{t-1,\theta},\; \sigma_t^2 |\Delta t|\, \mathbf{I}\bigr),
\end{equation}
which has a tractable log-probability, enabling the computation of the importance ratio $\rho_t^k(\theta)$ required by GRPO.

\subsection{Closed-Form KL Divergence}
\label{sec:supp_kl}

Since both $\pi_\theta(\mathbf{x}_{t-1} \mid \mathbf{x}_t, p)$ and $\pi_{\mathrm{ref}}(\mathbf{x}_{t-1} \mid \mathbf{x}_t, p)$ are isotropic Gaussians with the same variance $\sigma_t^2 |\Delta t|$ (only the means differ), the per-step KL divergence admits a simple closed form:
\begin{equation}
\label{eq:supp_kl}
D_{\mathrm{KL}}\bigl(\pi_\theta(\cdot \mid \mathbf{x}_t, p) \,\|\, \pi_{\mathrm{ref}}(\cdot \mid \mathbf{x}_t, p)\bigr)
= \frac{\|\bar{\mathbf{x}}_{t-1,\theta} - \bar{\mathbf{x}}_{t-1,\mathrm{ref}}\|^2}{2\,\sigma_t^2\,|\Delta t|}.
\end{equation}
Substituting the Euler--Maruyama means, this reduces to a scaled squared difference between the velocity predictions:
\begin{equation}
\label{eq:supp_kl_velocity}
D_{\mathrm{KL}}\bigl(\pi_\theta \,\|\, \pi_{\mathrm{ref}}\bigr)
= \frac{|\Delta t|}{2}\left(\frac{\sigma_t(1-t)}{2t} + \frac{1}{\sigma_t}\right)^2 \|v_\theta(\mathbf{x}_t, t, p) - v_{\mathrm{ref}}(\mathbf{x}_t, t, p)\|^2.
\end{equation}
This closed-form expression avoids costly sampling-based KL estimates and provides a smooth, well-behaved regularization signal at every denoising step.

\subsection{Denoising Reduction}
\label{sec:supp_denred}

Online RL requires repeatedly generating samples from the current policy to collect training data.
For flow models, each sample requires $T$ sequential denoising steps, making data collection the dominant computational bottleneck, especially for large video models.

Flow-GRPO~\citep{flowgrpo} introduces a \emph{denoising reduction} strategy to address this.
During training, samples are generated with a significantly reduced number of denoising steps $T_{\mathrm{train}} \ll T_{\mathrm{infer}}$ (\eg, $T_{\mathrm{train}} = 10$ vs.\ $T_{\mathrm{infer}} = 40$).
Although the resulting training samples are of lower visual quality than those produced with the full schedule, they still carry informative reward signals: the reward function can distinguish better from worse samples even at reduced quality.
At inference time, the full denoising schedule is restored, and the policy improvements learned from the reduced-step training transfer effectively to the high-quality generation regime.

This strategy provides a substantial speedup (proportional to $T_{\mathrm{infer}} / T_{\mathrm{train}}$) in the online RL data collection loop without degrading the final sample quality at test time, as verified empirically in~\citet{flowgrpo}.

\section{More Experiment Results}
\label{sec:supp_training}
\subsection{More Training Details}
\noindent\textbf{Latent geometry model.}
We train the latent geometry model for 20 epochs using the AdamW optimizer with a learning rate of $2\times10^{-4}$ and no weight decay. 
Gradient clipping is applied with a maximum norm of 1.0. 
The learning rate is scheduled with cosine decay and a linear warmup over the first 100 optimization steps. 
The stitching connector is implemented as a 3D convolutional layer with kernel size $5\times5\times5$, stride $1\times2\times2$, and padding $2\times2\times2$. 
For LoRA adaptation~\citep{lora}, we set the rank to $r=64$ and the scaling factor to $\alpha=32$.

\noindent\textbf{VGGRPO.}
For VGGRPO training, we optimize the model with AdamW using a learning rate of $1\times10^{-4}$ and weight decay of $1\times10^{-4}$. 
The overall training requires approximately 1536 GPU hours.
Unless otherwise stated, LoRA~\citep{lora} is applied with rank $r=32$ and scaling factor $\alpha=64$, and gradient clipping is performed with a maximum norm of 1.0.
For policy optimization, the clipping range $\varepsilon$ for the policy ratio $\rho_t^k(\theta)$ is set to $1\times10^{-3}$, while the KL regularization weight $\beta$ is fixed at 0.004.

\subsection{More Details of Test-Time Reward Guidance}
\label{app:guidance_impl}

Our latent geometry model is fully differentiable, which enables \emph{test-time reward guidance} directly in latent space.
Unlike RGB-space guidance, our approach does not require decoding intermediate latents into video frames.
Instead, at selected denoising steps, we back-propagate the gradient of the geometry reward through the latent geometry model and use it to adjust the diffusion update.

Given a noisy latent $\mathbf{z}_t$ at timestep $t$, the video diffusion model predicts a velocity field $v_\theta(\mathbf{z}_t, t, p)$ under text condition $p$.
We then feed $\mathbf{z}_t$ into the latent geometry model $\hat{\mathrm{\Phi}}_\psi$ to obtain the geometric quantities required for reward computation.
Based on these predictions, we define the latent-space reward as
\begin{equation}
r(\mathbf{z}_t)
=
\lambda_{\text{motion}} \, r_{\mathrm{motion}}(\mathbf{z}_t)
+
\lambda_{\text{geo}} \, r_{\mathrm{geo}}(\mathbf{z}_t),
\end{equation}
where $\lambda_{\text{motion}}$ and $\lambda_{\text{geo}}$ denote the weights of the camera motion smoothness reward and the geometry reprojection consistency reward, respectively.

Following~\cite{liuimproving}, we compute the gradient of this reward with respect to the current latent and use it to modify the sampling trajectory:
\begin{equation}
\tilde{v}_\theta(\mathbf{z}_t, t, p)
=
v_\theta(\mathbf{z}_t, t, p)
-
s_{\text{reward}} \frac{t}{1-t}
\nabla_{\mathbf{z}_t} r(\mathbf{z}_t),
\end{equation}
where $s_{\text{reward}}$ controls the strength of reward guidance.
The gradient is obtained by back-propagating through the reward computation and the latent geometry model, entirely in latent space and without RGB decoding.

In practice, we combine this reward guidance with classifier-free guidance in the standard way.
To limit the additional inference cost, we do not apply reward guidance at every denoising step.
Unless otherwise specified, we use 50 denoising steps in total and apply reward guidance once every 20 steps.
This sparse guidance already yields consistent improvements in geometric consistency, as shown in \Cref{tab:ab_guidance}, while keeping the runtime overhead modest.
Pseudo-code is provided in \Cref{alg:reward_guidance}, illustrating a simple training-free way to improve world consistency at inference time.

\vspace{8pt}
\begin{lstlisting}[style=pythonstyle, caption={PyTorch-style pseudo-code for latent-space test-time reward guidance using our latent geometry model.}, label={alg:reward_guidance}]
def latent_reward_guidance(
    model,
    latent_geometry_model,
    latents,
    prompt_embeds,
    timesteps,
    reward_guidance_scale,
    reward_weights,
    cfg_scale=1.0,
    guidance_interval=20,
):
    """
    model: video diffusion backbone predicting velocity
    latent_geometry_model: differentiable latent geometry model
    latents: initial noisy video latents
    prompt_embeds: text condition
    timesteps: sampling timesteps
    reward_guidance_scale: strength of reward guidance
    reward_weights: weights for different rewards
    cfg_scale: classifier-free guidance scale
    guidance_interval: apply reward guidance once every k steps
    """

    dts = timesteps[:-1] - timesteps[1:]

    for i, t in enumerate(timesteps[:-1]):
        # Standard conditional prediction
        v_pred = model(latents, prompt_embeds, t)

        # Classifier-free guidance
        if cfg_scale != 1.0:
            v_uncond = model(latents, None, t)
            v_pred = v_uncond + cfg_scale * (v_pred - v_uncond)

        # Sparse reward guidance
        use_guidance = ((i + 1) % guidance_interval == 0)

        if use_guidance and t < 1:
            latents = latents.detach().requires_grad_(True)

            # Decode geometry directly from latent space
            geom_outputs = latent_geometry_model(latents)

            # Compute geometry-aware reward
            reward_smooth = camera_motion_smoothness_reward(geom_outputs)
            reward_geo = geometry_reprojection_reward(geom_outputs)

            reward = (
                reward_weights["smooth"] * reward_smooth +
                reward_weights["geo"] * reward_geo
            )

            # Gradient of reward w.r.t. current noisy latents
            grad = torch.autograd.grad(reward, latents)[0]

            # Modify velocity field
            v_pred = v_pred - reward_guidance_scale * t / (1 - t) * grad

        # Solver update
        latents = latents - dts[i] * v_pred

    return latents
\end{lstlisting}

\begin{figure*}[t] 
    \centering
    \includegraphics[width=\textwidth]{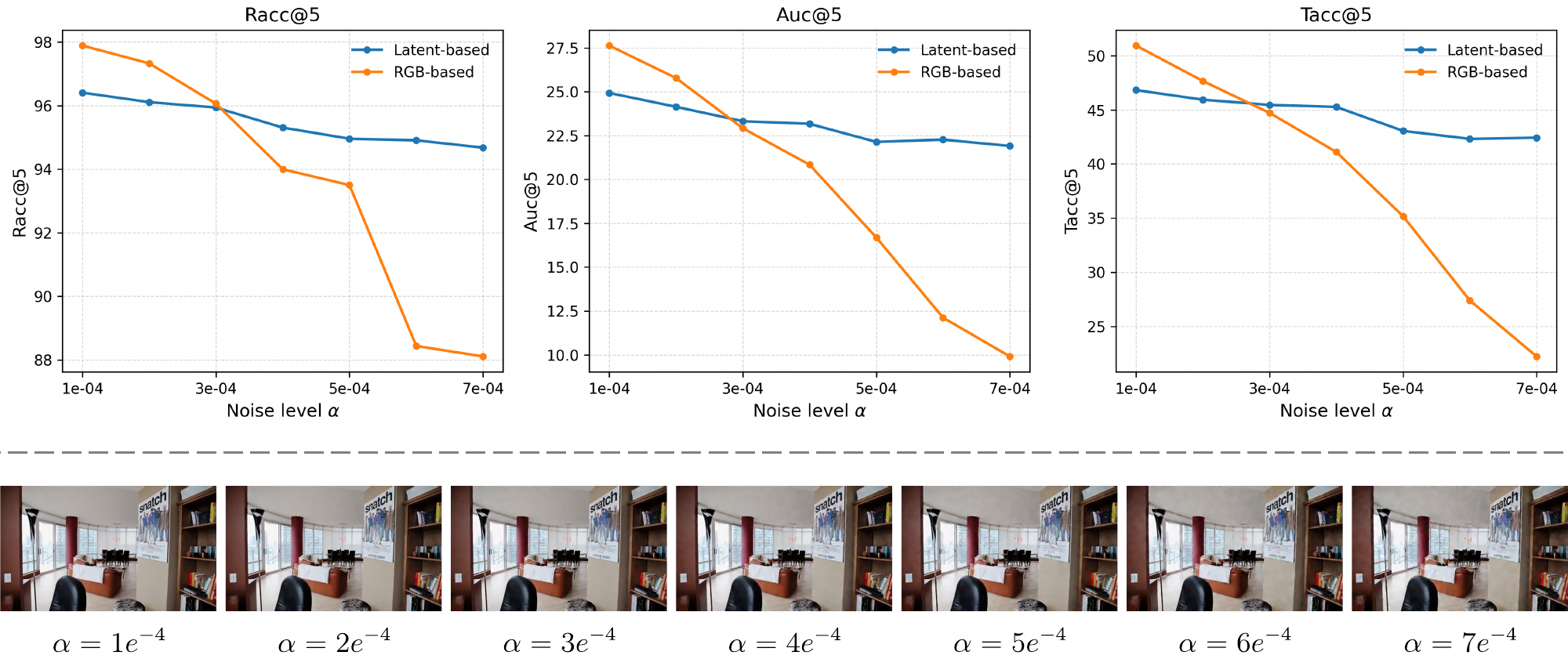} 
    \caption{
    \textbf{Analysis of the Latent Geometry Model.}
    We compare the latent geometry model with the original RGB-based geometry model on 50 RealEstate10K test sequences under controlled perturbations applied in the video latent space.
    The top row reports camera pose estimation performance as the perturbation scale $\alpha$ increases, measured by relative rotation accuracy (\textit{left}, Racc@5), area under the accuracy curve (\textit{middle}, AUC@5), and relative translation accuracy (\textit{right}, Tacc@5).
    Our latent geometry model maintains stable performance across all noise levels, whereas the RGB-based geometry model degrades substantially as the perturbation grows.
    The bottom row shows a decoded RGB frame from the perturbed latents at different values of $\alpha$.
    Even when perturbations produce only barely perceptible visual changes in RGB space, the RGB-based geometry model already degrades, reflecting the distribution gap when applied to generated content rather than real images. Our latent geometry model, trained directly on generated latents, avoids this gap and remains robust.
    }
    \label{fig:supp_plot} 
\end{figure*}

\subsection{More Quantitative Comparisons}
\noindent\textbf{Latent geometry model.}
To further examine the ability of the latent geometry model, we compare our stitched latent geometry model with the original RGB-based geometry model~\citep{karhade2025any4d} on 50 test video sequences from RealEstate10K~\citep{re10k} under controlled perturbations in the latent space.
Specifically, given a video latent representation $\mathbf{z}$, we inject Gaussian noise as
\begin{equation}
    \mathbf{z}' = \mathbf{z} + \alpha \, \|\mathbf{z}\| \, \epsilon, \qquad \epsilon \sim \mathcal{N}(0, I),
\end{equation}
where $\alpha$ controls the perturbation strength. 
This setup emulates a challenge in RGB-based reward evaluation: the geometry model is pretrained on real images, yet at reward computation time it receives decoded frames from a diffusion model, which differ from its training distribution.
For the original geometry model, we decode the perturbed latent $\mathbf{z}'$ back into RGB frames and use them as input.
For our latent geometry model, the perturbed latent is directly used as input without decoding.

We evaluate relative camera pose estimation using three standard metrics: rotation accuracy (Racc@5), translation accuracy (Tacc@5), and area under the accuracy curve (AUC@5), where ``@5'' denotes evaluation under a 5$^\circ$ error threshold.
Here, AUC@5 summarizes the cumulative accuracy of the joint relative pose error up to 5$^\circ$.
As shown in \Cref{fig:supp_plot}, the original RGB-based geometry model is highly sensitive to latent perturbations, reflecting the distribution gap when applied to generated rather than real content: performance degrades rapidly as $\alpha$ increases, even when the decoded RGB frames remain visually almost unchanged, as illustrated in the bottom row of \Cref{fig:supp_plot}.
By contrast, our latent geometry model remains considerably more stable across all perturbation levels, exhibiting only modest degradation.
These results show that decoding latents to RGB before geometry evaluation is highly sensitive to distribution shifts between real and generated content. Our latent geometry model, trained directly on video latents, closes this domain gap and provides substantially more reliable geometry estimates. This robustness supports the use of latent-space rewards in VGGRPO: unlike RGB-based rewards, latent rewards remain stable under the unavoidable noise and distribution shifts in generated video.

\begin{table*}[tb]
  \footnotesize
  \centering
  \renewcommand{\arraystretch}{1}
    \centering
    \resizebox{1\linewidth}{!}{%
        \setlength{\tabcolsep}{3pt}
        \renewcommand{\arraystretch}{1.26}
        \begin{tabular}{lcccccc}
          \toprule
          
      \textbf{Model}
      & \textbf{Sub. Cons.}\,$\uparrow$
      & \textbf{Bg. Cons.}\,$\uparrow$
      & \textbf{Aes. Qual.}\,$\uparrow$ 
      & \textbf{Img. Qual.}\,$\uparrow$
      & \textbf{Mot. Smooth.}\,$\uparrow$
      & \textbf{Dyn. Deg.}\,$\uparrow$  \\
          \midrule
            Baseline             &  0.9542 & 0.9528  & 0.5966 & \underline{0.6733}  & 0.9841 & \textbf{0.4237}  \\
            SFT                  &  0.9578 & 0.9556  & 0.5968 & 0.6304  & \underline{0.9872} & \underline{0.3988}  \\
            Epipolar-DPO         &  0.9601 & 0.9564  & \underline{0.5977} & 0.6353  & 0.9847 & 0.3325  \\
            VideoGPA             &  \underline{0.9605} & \underline{0.9565}  & 0.5973 & 0.6338  & 0.9835 & 0.3419  \\
           \rowcolor{row} Ours   &  \textbf{0.9644} & \textbf{0.9583}  & \textbf{0.5991} & \textbf{0.6861}  & \textbf{0.9895} & 0.3962  \\
          \bottomrule
        \end{tabular}
    }
  \vspace{0.1in}
\caption{\textbf{Generalization Performance on Standard VBench Captions.}
Compared with the base model, prior post-training methods provide only limited gains and even degrade certain metrics, such as \textit{Imaging Quality}.
In contrast, VGGRPO achieves the best overall performance on most metrics among the base model and prior post-training baselines.
These results show that VGGRPO transfers effectively to the standard VBench caption set, improving generation quality while maintaining robust generalization across diverse scenarios.
The best and second-best results are highlighted in \textbf{bold} and \underline{underlined}, respectively.
}
\label{tab:supp_vbench}
\end{table*}

\noindent\textbf{Generalization performance.}
Complementing~\Cref{tab:ab_vbench} in the main paper, we further compare VGGRPO with additional post-training baselines on the standard VBench caption set to provide a direct evaluation of different post-training strategies.
As shown in \Cref{tab:supp_vbench}, these post-training baselines yield only limited overall gains in general video quality.
In particular, all prior post-training baselines score lower than the base model on the \textit{Imaging Quality} metric, suggesting that these methods potentially compromise the broad generative quality and generalization ability inherited from large-scale pre-training.

In contrast, VGGRPO not only improves world consistency, but also achieves the strongest overall performance across most metrics on the standard VBench caption set, consistently outperforming both the base model and prior post-training methods.
These results indicate that our geometry-aware post-training preserves, rather than sacrifices, the general-purpose video generation quality of the pre-trained model.
Dynamic Degree is driven by optical flow magnitude~\citep{teed2020raft}, which means that unstable camera trajectories and abrupt scene changes artificially inflate the metric. As shown in our webpage videos, the baseline often produces excessively fast, jittery trajectories, which raise the score. VGGRPO produces smoother, more stable motion, naturally reducing the flow-based metric while still maintaining a much smaller gap to the baseline than prior post-training methods.
Overall, these results show that VGGRPO improves world-consistent video generation while maintaining strong perceptual quality and robust generalization across diverse scenarios.

\subsection{More Qualitative Results}

World-consistent video generation is important for many downstream applications. 
However, current models still frequently suffer from unstable camera motion and inconsistent geometry structure, leading to failures of world consistency in generated videos. 
VGGRPO provides an effective way to address both issues jointly in both static and dynamic scenes, resulting in more temporally stable and geometrically consistent video generation.

We provide additional qualitative comparisons in our \projectpage.
These examples cover a broad range of scenarios, including indoor environments, outdoor views, static scenes, and dynamic scenes with complex object motion.
Across these diverse cases, the baselines often exhibit characteristic failure modes such as camera shake, motion blur, geometric distortion, and broken temporal continuity.
By contrast, VGGRPO consistently produces smoother camera trajectories and more stable scene geometry, yielding more world-consistent videos.

These qualitative results also highlight the flexibility and effectiveness of VGGRPO beyond static settings, in contrast to prior post-training methods, which are largely restricted to static scenarios.
In challenging dynamic examples, the baselines suffer from severe scene blur and substantial geometric deformation, whereas VGGRPO preserves coherent scene structure and temporal continuity throughout the sequence.
Overall, these additional visualizations further demonstrate that VGGRPO effectively improves both camera stability and cross-view geometric coherence, leading to more realistic and world-consistent video generation across diverse scenarios.

\end{document}